\pdfoutput=1

\documentclass[11pt]{article}

\usepackage[final]{acl}
\usepackage{times}
\usepackage{latexsym}

\usepackage[T1]{fontenc}

\usepackage[utf8]{inputenc}

\usepackage{microtype}

\usepackage{inconsolata}

\usepackage{graphicx}

\usepackage{amsfonts}
\usepackage{amsmath}
\usepackage{amssymb}
\usepackage[ruled,linesnumbered]{algorithm2e}
\usepackage{booktabs}
\usepackage{diagbox}
\usepackage{enumerate}
\usepackage{hyperref}
\usepackage{makecell}
\usepackage{mathrsfs}
\usepackage{mathtools}
\usepackage{multirow}
\usepackage{soul}
\usepackage{subfigure}
\usepackage{subcaption}
\usepackage[most]{tcolorbox}
\tcbuselibrary{breakable}
\usepackage{textgreek}
\usepackage{url}
\usepackage{wrapfig}

\newcommand{\dif}{\mathrm{d}}

\definecolor{lightblue}{RGB}{165, 201, 235}
\sethlcolor{lightblue}

\title{InvestAlign: Overcoming Data Scarcity in Aligning Large Language Models with Investor Decision-Making Processes under Herd Behavior}

\author{\textbf{Huisheng Wang\thanks{Equal contribution.}\quad Zhuoshi Pan$^*$\quad Hangjing Zhang$^*$} \\
  \textbf{Mingxiao Liu$^*$\quad Hanqing Gao\quad H. Vicky Zhao\thanks{Corresponding author.}} \\
  Department of Automation, Tsinghua University, Beijing, China \\
  \texttt{\{whs22,pzs23,hangjing23,mx-liu21,gaohq22\}@mails.tsinghua.edu.cn}\\ \texttt{vzhao@tsinghua.edu.cn} \\}

\begin{document}
\maketitle
\begin{abstract}
Aligning Large Language Models (LLMs) with investor decision-making processes under herd behavior is a critical challenge in behavioral finance, which grapples with a fundamental limitation: the scarcity of real-user data needed for Supervised Fine-Tuning (SFT). While SFT can bridge the gap between LLM outputs and human behavioral patterns, its reliance on massive authentic data imposes substantial collection costs and privacy risks. We propose \textbf{InvestAlign}, a novel framework that constructs high-quality SFT datasets by leveraging theoretical solutions to similar and simple optimal investment problems rather than complex scenarios. Our theoretical analysis demonstrates that training LLMs with \textbf{InvestAlign}-generated data achieves faster parameter convergence than using real-user data, suggesting superior learning efficiency. Furthermore, we develop \textbf{InvestAgent}, an LLM agent fine-tuned with \textbf{InvestAlign}, which demonstrates significantly closer alignment to real-user data than pre-SFT models in both simple and complex investment problems. This highlights our proposed \textbf{InvestAlign} as a promising approach with the potential to address complex optimal investment problems and align LLMs with investor decision-making processes under herd behavior. Our code is publicly available at \url{https://github.com/thu-social-network-research-group/InvestAlign}.
\end{abstract}

\section{Introduction}
\label{sec: introduction}
In financial markets, investors typically make decisions based on their risk preferences to achieve higher returns, lower volatility, and maximize their utility. \cite{merton1969lifetime}. Investment decisions are crucial as they not only impact individual financial outcomes but also shape market dynamics and overall economic stability, making them a key driver of both personal wealth and broader market efficiency \cite{ahmad2022does}.
During this process, investment assistants such as financial analysts and fund managers, play a significant role by sharing their own investment decisions through platforms \cite{brown2008neighbors}. These investment assistants often have rich investment experience and extensive influence, leading investors to mimic their behaviors. This is commonly referred to as herd behavior in microeconomics and behavioral finance \cite{bikhchandani2000herd}. The prior works in \cite{wang2024optimal,wang2024herd,wang2025optimal} have investigated the optimal investment problem considering herd behaviors between two agents, and theoretically analyzed the impact of herd behavior on their decisions. However, there are more complex problems where the above models fall short or only provide qualitative insights \cite{zhou2022internet}, prompting us to explore alternative approaches.

Large Language Models (LLMs) have been widely adopted in various domains as generative agents to assist with specific tasks \cite{kovavc2023socialai,m2024augmenting}. A notable trend is the enhancement of LLM agents with human-like intelligence to simulate human decision-making processes \cite{gao2024large}. In economics and finance, substantial works have been done on aligning LLMs with human values and decisions, particularly in models for market behavior prediction and the analysis of complex economic data for policy-making \cite{zhao2023survey,lee2024survey}. 
These studies predominantly address macroeconomic concerns, such as the dynamics of information dissemination and collective decision-making within global markets \cite{li2024econagent}. To our best knowledge, there has been limited exploration of LLMs' efficacy in microeconomics and behavioral finance, and current LLMs do not fully align with investor decision-making processes, as shown in Section \ref{sec:formulation}. 

\begin{figure*}[!t]
\centering
\includegraphics[width=\linewidth]{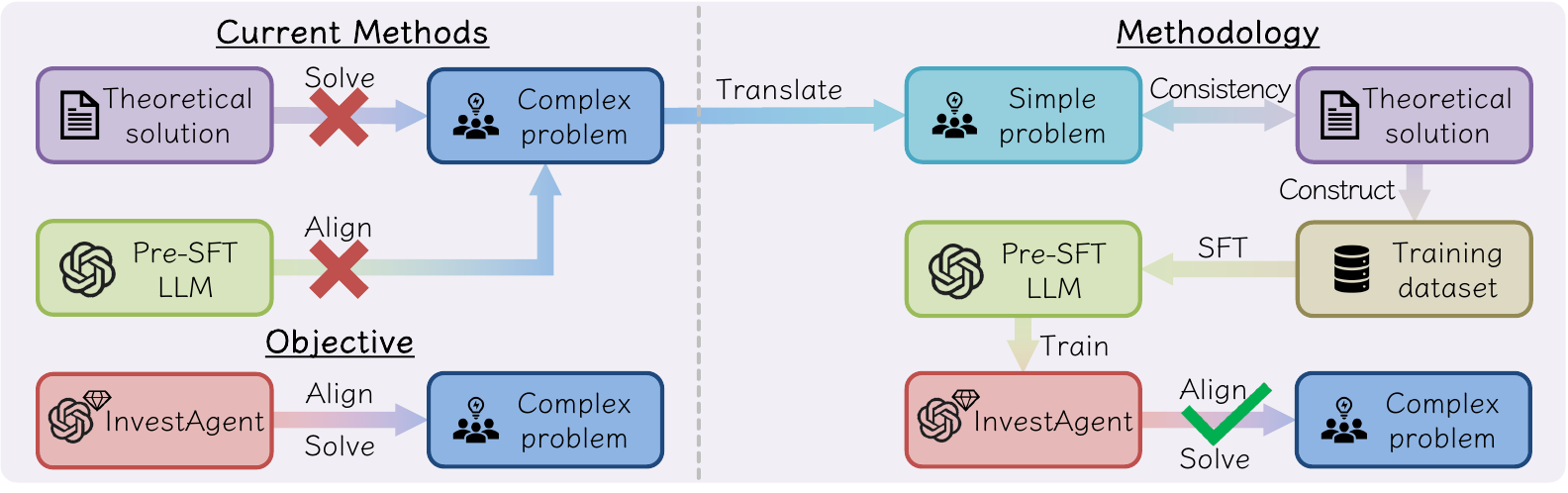}
\caption{Overview of \textbf{InvestAlign}.}
\label{fig: investalign}
\vspace{-3mm}
\end{figure*}

Achieving the alignment of LLMs to investor decision-making processes often relies on large-scale real-user data in Supervised Fine-Tuning (SFT) \cite{zhang2023instruction}. Fine-tuned with specific training datasets, LLMs can better generate investor behavior in complex problems. However, it faces the following obstacles. Collecting real-user data can be costly due to the wide variation in investment attributes like risk preference and herd degree \cite{abbot2017heterogeneous}. Additionally, many investors are reluctant to share their investment decisions due to privacy and security concerns.

To address data scarcity, note that for some simple problems such as the one in \cite{wang2024herd}, the theoretical solution has already been found, using which we can generate a large amount of training data. One possible solution is that, given a complex problem, we first identify a similar and simple problem with a theoretical solution, construct the SFT dataset using this theoretical solution, and then fine-tune LLMs to solve the original complex problem. There are several issues to be addressed when following this approach:

\noindent$\bullet\ \mathsf{Q_1}$: Given the complex problem, how to identify a similar and simple problem?

\noindent$\bullet\ \mathsf{Q_2}$: Do the theoretical solutions of the simple problem align with real users' investment decisions, and can they be used to construct a training dataset that mirrors investor decision-making processes?

\noindent$\bullet\ \mathsf{Q_3}$: How can we generate the training dataset based on the theoretical solution of the simple problem? How does it align with investor decision-making processes compared with real-user data?

\noindent$\bullet\ \mathsf{Q_4}$: How to adapt the fine-tuned LLMs to solve the complex problem, and what is its performance?

To validate the feasibility of the proposed approach and address the above four issues, in this work, we examine an optimal investment scenario involving two agents as a case study. We consider the following two primary factors influencing herd behavior. The first factor is the pattern of herd behavior, which includes \textit{absolute herd behavior} in \cite{wang2024herd}, where agents replicate the entire portfolio of others, and \textit{relative herd behavior} in \cite{wang2024optimal}, where agents mimic the changing rate of others' decisions. The second factor is the structure of the influence network, which includes \textit{unilateral influence} from one agent to another and \textit{mutual influence} between two agents \cite{wang2025optimal}. We investigate two complex problems corresponding to relative herd behavior under unilateral influence and absolute herd behavior under mutual influence, respectively. Although theoretical solutions for these problems exist, their computational complexity is notably high. To address $\mathsf{Q_1}$, we utilize absolute herd behavior under unilateral influence as the simple problem, for which the theoretical solution is more readily derived. Note that while the simple problem shares mathematical similarities with the original complex problems, they differ in their approaches to measuring herd behavior.

To answer $\mathsf{Q_2}$, we collect real-user data on the simple problem and apply statistical methods to validate the consistency between real-user data and the theoretical solution. Next, to answer $\mathsf{Q_3}$, we construct SFT datasets based on the theoretical solutions, and theoretically prove that fine-tuning LLMs on the above training datasets leads to faster parameter convergence than using real-user data. Then, to answer $\mathsf{Q_4}$, given the training dataset, we fine-tune the LLMs and develop the \textbf{InvestAgent}s, which can make decisions similar to the theoretical solution, thus aligning with real-user data in the simple problem. Finally, we conduct another real-user test to verify the performance of \textbf{InvestAgent}s on solving the original complex problems, and experimental results show that \textbf{InvestAgent}s exhibit better alignment performance than pre-SFT LLMs.

In conclusion, our contributions include: 

\noindent$\bullet$ We explore and utilize LLMs in micro economics and behavioral finance, particularly in the domain of optimal investment under herd behavior.

\noindent$\bullet$ We effectively construct a large amount of high-quality training data with the theoretical solution of the corresponding mathematical model.

\noindent$\bullet$ We propose the LLM alignment techniques, \textbf{InvestAlign}, using the generated abundant dataset and apply SFT to fine-tune LLMs.

\section{Related Work}
\label{sec:related}

\textbf{LLMs in Finance and Optimal Investment.}\;
For finance-related tasks, several specialized LLMs have been developed, e.g., BloombergGPT \cite{wu2023bloomberggpt}, FinGPT \cite{yang2023fingpt}, and XuanYuan \cite{zhang2023xuanyuan}. The success of these models depends on large amounts of training data, and the challenge is how to effectively collect and generate high-quality data, which is a key goal of our proposed method. Focusing on the optimal investment problem, prior studies have explored the use of LLMs in different scenarios such as investment idea generation and quantitative investment \cite{li2023multimodal,wang2023alpha,yu2024finmem}. Multi-agent frameworks incorporating LLM-based trading strategies combined with quantitative methods \cite{kou2024automate} have demonstrated superior Sharpe ratios and lower maximum drawdowns. Other approaches model real trading firms \cite{xiao2024tradingagents} or use cognitive-inspired multi-agent systems to replicate human decision hierarchies \cite{li2023tradinggpt}. However, within agent-based modeling, only a few works use LLMs as generative investors to simulate or complement human investor behavior, e.g., InvestLM in \cite{yang2023investlm}, EconAgent in \cite{li2024econagent}, and StockAgent in \cite{zhang2024ai}. Similar agent-based ideas have been widely used in many areas, such as problems in the economic system \cite{horton2023large,chen2023emergence,geerling2023chatgpt}, social science \cite{ghaffarzadegan2023generative,liu2024exploring,wang2024evaluating}, and natural science \cite{boiko2023emergent,m2024augmenting}. While several studies in other domains have explored the LLMs' irrational behaviors to mirror human cognitive biases \cite{liu2024exploring,wang2024behavioral,xiao2024behavioral}, existing agent-based LLMs for investment have not yet accounted for the herd behavior \cite{bikhchandani2000herd}, which is significant in microeconomics and behavioral finance. Understanding its influence on the optimal investment problem while incorporating LLMs is crucial for analyzing investor behavior \cite{ahmad2022does}. 

\noindent\textbf{LLM Alignment.}\;
LLM alignment with human values has emerged as a critical area of research \cite{wang2024comprehensive}, aiming to make LLM agents behave in line with human intentions and values \cite{ji2023ai}. Although LLMs excel in various tasks, issues like untruthful answers \cite{bang2023multitask}, sycophancy \cite{perez2022discovering}, and deception \cite{steinhardt2023emergent}, raise concerns about controllability and risks in LLM agents. To achieve forward alignment, which ensures that trained systems meet alignment requirements, numerous methods for policy learning and scalable oversight are proposed \cite{ji2023ai,wang2024comprehensive}. For LLMs, a typical approach is reinforcement learning from human feedback (RLHF) using SFT \cite{christiano2017deep,bai2022training,bowman2022measuring,wang2024unifying}. In microeconomics and behavioral finance, only a limited number of studies involve LLMs \cite{li2024econagent,horton2023large}, focusing on macro-level alignment while ignoring microcosmic behaviors of human decision-making.

\noindent\textbf{SFT Methods in Optimal Investment.}\;
SFT is a widely adopted technique in the field of LLMs for improving model performance on specific tasks by refining pre-trained models with a dataset tailored to the target task \cite{zhang2023instruction}. Many tricks and methods of SFT have been proposed to achieve better LLM alignment to humans, e.g., \cite{ding2023enhancing,wang2023far,xie2024minor,li2024getting}. In the domain of finance, SFT has been applied to various investment-related tasks such as price prediction, financial reports summarization, sentiment analysis, portfolio optimization, etc. \cite{zhao2024snfinllm,guo2024fine,an2024finverse}. These advancements highlight the power of SFT in tailoring LLMs to meet the specific needs of investment strategies, enabling models to simulate or complement human-like behaviors. However, collecting large, high-quality datasets for fine-tuning in optimal investment remains a challenging problem \cite{abbot2017heterogeneous}. Existing research explores synthetic financial data generation using generative adversarial networks to replicate complex properties, such as stock prices and trading volumes \cite{assefa2020generating,ramzan2024generative}. However, these methods face limitations due to data scarcity and regulatory constraints \cite{potluru2023synthetic}. Privacy-preserving federated learning addresses these challenges by enabling secure cross-institutional data generation \cite{behera2022fedsyn,chen2022importance}. Building on this, the work in \cite{balch2024six} introduces a hierarchical privacy framework to evaluating synthetic data risks in financial institutions. However, these approaches often fail to capture the nuanced behaviors in real-world investment decisions.

\section{Problem Simplification \texorpdfstring{\&}{and} Real-User Data Verification}
\label{sec:formulation}
To verify the feasibility of the proposed method \textbf{InvestAlign}, we consider the optimal investment scenario involving two agents $\mathsf{A_1}$ and $\mathsf{A_2}$. As mentioned above, for the first complex problem $\mathsf{P_1}$, we assume that $\mathsf{A_1}$'s investment decisions are unilaterally influenced by $\mathsf{A_2}$ under relative herd behavior, and for the second problem $\mathsf{P_2}$, we assume that $\mathsf{A_1}$'s and $\mathsf{A_2}$'s investment decisions are mutually influenced under absolute herd behavior. For the simple problem with the theoretical solution, denoted by $\mathsf{P_3}$, we assume that $\mathsf{A_1}$'s investment decisions are unilaterally influenced by $\mathsf{A_2}$ under absolute herd behavior. Next, to answer $\mathsf{Q_2}$, we collect real-user data using interviews and questionnaires, and obtain pre-SFT LLMs' investment decisions for $\mathsf{P_3}$. Then, we show that pre-SFT LLMs' responses are misaligned with the real-user data, and validate the statistical consistency between the theoretical solutions and the real-user data. 

\subsection{Optimal Investment Problems}
\label{sec:problem}
Following the work in \cite{merton1969lifetime}, we consider the scenario where $\mathsf{A_1}$ and $\mathsf{A_2}$ invest in the period $\mathcal{T}=[0,T]$ in a financial market consisting of a deposit and a stock. We define $\mathsf{A}_i$'s fund invested in the stock as his/her \textit{investment decisions}, denoted by $\{P_i(t)\}_{t\in\mathcal{T}}$ ($i=1,2$). We denote $r$ as the interest rate of the deposit, and $v$ and $\sigma$ as the excess return rate and volatility of the stock. Given the above parameters, $\mathsf{A}_i$'s fund $\{X_i(t)\}_{t\in\mathcal{T}}$ satisfies
\begin{equation}\textstyle
    \dif X_i(t)\!=\![rX_i(t)+vP_i(t)]\dif t+\sigma P_i(t)\dif W(t),\!\!
\label{eq:budget}
\end{equation}
where $X_i(0)=x_{i,0}$ is his/her initial fund, and $\{W(t)\}_{t\in\mathcal{T}}$ is a standard Brownian motion modeling the randomness of the stock price. Considering the herd behavior, $\mathsf{A}_i$ jointly maximizes his/her expected utility of the terminal fund $\mathbb{E}\phi_i[X_i(T)]$ and minimizes the distance between his/her own and the other's decisions $D(P_1,P_2)$. Following the work in \cite{rogers2013optimal}, we assume that $\mathsf{A}_i$'s utility is $\phi_i[X_i(T)]=-\frac{1}{\alpha_i}\exp[-\alpha_iX_i(T)]$, where $\alpha_i$ is his/her risk aversion coefficient. In summary, the general optimal investment problem is
\begin{equation}\left\{
    \begin{aligned}
        &\textstyle
        \sup_{\{P_1(t)\}_{t\in\mathcal{T}}}\mathbb{E}\phi_1[X_1(T)]-\theta_1D(P_1,P_2),\\
        &\textstyle
        \sup_{\{P_2(t)\}_{t\in\mathcal{T}}}\mathbb{E}\phi_2[X_2(T)]-\theta_2D(P_1,P_2),
    \end{aligned}\right.\!\!\!
    \label{eq:optimal}
\end{equation}
where $\theta_i$ is $\mathsf{A}_i$'s influence coefficient to address the tradeoff between the two different objectives. We define the risk aversion coefficient $\alpha_i$ and the influence coefficient $\theta_i$ as $\mathsf{A}_i$'s \textit{investment attribute}. 

\noindent$\bullet\ \mathsf{P_1}$\textbf{: Optimal investment problem under relative herd behavior with unilateral influence.}
Following the work in \cite{wang2024optimal}, when considering the relative herd behavior, the distance is defined as $\delta(P_1,P_2)=\frac{1}{2}\int_\mathcal{T}[P_1'(t)-P_2'(t)]^2\dif t$, i.e., the integrated square error between the two decisions' changing rates, and when considering the unilateral influence of $\mathsf{A_2}$ on $\mathsf{A_1}$, $\mathsf{A_2}$'s influence coefficient $\theta_2=0$. In this case, the optimal investment problem (\ref{eq:optimal}) becomes $\mathsf{P_1}$, which is
\begin{equation}\left\{
    \begin{aligned}
        &\textstyle
        \sup_{\{P_1(t)\}_{t\in\mathcal{T}}}\mathbb{E}\phi_1[X_1(T)]-\theta_1\delta(P_1,P_2),\\
        &\textstyle
        \sup_{\{P_2(t)\}_{t\in\mathcal{T}}}\mathbb{E}\phi_2[X_2(T)].
    \end{aligned}\right.\!\!
    \label{equ: problem_relative}
\end{equation}

\noindent$\bullet\ \mathsf{P_2}$\textbf{: Optimal investment problem under absolute herd behavior with mutual influence.}
Following the work in \cite{wang2025optimal}, when considering the absolute herd behavior, the distance is defined as $\varDelta(P_1,P_2)=\frac{1}{2}\int_\mathcal{T}[P_1(t)-P_2(t)]^2\dif t$, i.e., the integrated square error between the two agents' decisions, and when considering the mutual influence, the two agents' influence coefficients, $\theta_1$ and $\theta_2$, are both positive. In this case, the optimal investment problem (\ref{eq:optimal}) becomes $\mathsf{P_2}$, which is
\begin{equation}\left\{
    \begin{aligned}
        &\textstyle
        \sup_{\{P_1(t)\}_{t\in\mathcal{T}}}\mathbb{E}\phi_1[X_1(T)]-\theta_1\varDelta(P_1,P_2),\\
        &\textstyle
        \sup_{\{P_2(t)\}_{t\in\mathcal{T}}}\mathbb{E}\phi_2[X_2(T)]-\theta_2\varDelta(P_1,P_2).
    \end{aligned}\right.\!\!
    \label{equ: problem_mutual}
\end{equation}

\noindent$\bullet\ \!\mathsf{P_3}$\textbf{:\! Optimal investment problem under absolute herd behavior with unilateral influence} is
\begin{equation}\left\{
    \begin{aligned}
        &\textstyle
        \sup_{\{P_1(t)\}_{t\in\mathcal{T}}}\mathbb{E}\phi_1[X_1(T)]-\theta_1\varDelta(P_1,P_2),\\
        &\textstyle
        \sup_{\{P_2(t)\}_{t\in\mathcal{T}}}\mathbb{E}\phi_2[X_2(T)].
    \end{aligned}\right.\!\!
    \label{equ: problem-absolute}
\end{equation}
From the work in \cite{wang2024herd}, $\mathsf{A}_i$'s theoretical optimal decision, denote by $\{\hat{P}_i(t)\}_{t\in\mathcal{T}}$, of $\mathsf{P_3}$ can be easily calculated, as shown in (\ref{eq:theoretical}) in Appendix \ref{app:theory}. We set the parameter values of $\mathsf{P_1}$, $\mathsf{P_2}$, and $\mathsf{P}_3$ in Appendix \ref{sec:para}. 

\subsection{Data Collection}
\label{sec: data}

\textbf{Real-User Data Collection.}
To verify whether $\mathsf{P_3}$'s theoretical solution in (\ref{eq:theoretical}) matches users' investment decisions, we collect real-user data from $119$ participants using interviews and questionnaires when facing the investment problem $\mathsf{P_3}$. We denote the index set of participants as $\mathcal{I}=\{1,2,\ldots,119\}$. To reduce bias and noise in the collected data, we primarily recruit professionals and students in the fields of microeconomics and behavioral finance, and we treat the real-user data as a proxy for the ground truth. We let the participant play the role of $\mathsf{A_1}$ unilaterally influenced by an investment assistant $\mathsf{A_2}$ whose investment attribute is set in Appendix \ref{sec:para}.

The questionnaire we use is in Figure \ref{fig:questionnaire_absolute} in Appendix \ref{sec:questionnaire}. In the first part, we provide the task description, including the asset information and the participants' goals. In the second part, participants report their investment decisions, denoted by $\{\tilde{P}_1^i(t)\}_{t\in\mathcal{T}}$ for all $i\in\mathcal{I}$. To facilitate participants' decision-making, we ask them to report the proportions of funds invested in the stock $\{\tilde{P}_1^i(t)/X_1^i(t)\}_{t\in\mathcal{T}}$. When processing the data, we first calculate $\{X_1^i(t)\}_{t\in\mathcal{T}}$ using (\ref{eq:budget}), and then calculate the participants' decisions $\{\tilde{P}_1^i(t)\}_{t\in\mathcal{T}}$ according to the proportions $\{\tilde{P}_1^i(t)/X_1^i(t)\}_{t\in\mathcal{T}}$. 
In the third part, we ask the participants the information about their investment attributes, based on which, we calculate their risk aversion coefficients $\{\alpha_1^i\}_{i\in\mathcal{I}}$ and influence coefficients $\{\theta_1^i\}_{i\in\mathcal{I}}$. Details are in Appendix \ref{sec:attribute}. 

\noindent\textbf{Collection of Pre-SFT LLMs' Investment Decisions.}
Next, to verify whether pre-SFT LLMs align with real-user data, we collect the pre-SFT LLMs' investment decisions. We choose a variety of LLMs, including the API-based model \texttt{GPT-3.5-Turbo} \cite{achiam2023gpt}, as well as the open-source models including \texttt{Qwen-2-7B-Instruct} \cite{qwen2}, \texttt{Meta-Llama-3.1-8B-Instruct} \cite{dubey2024llama}, and \texttt{GLM-4-9B-CHAT}\footnote{The results of \texttt{GLM-4-9B-CHAT} are in Appendix \ref{app:glm}.} \cite{glm2024chatglm}. To obtain these pre-SFT LLMs' investment decisions in $\mathsf{P_2}$, we first construct a prompt, as shown in Figure \ref{fig:prompt_absolute} in Appendix \ref{sec:prompt}. The first part is identical to the questionnaire in Figure \ref{fig:questionnaire_absolute}, where we designate the pre-SFT LLM as an investment expert and describe the task. In the second part, we assign the pre-SFT LLM its investment attribute, corresponding to the participant's investment attribute $\{\alpha_1^i\}_{i\in\mathcal{I}}$ and $\{\theta_1^i\}_{i\in\mathcal{I}}$ in the real-user data. In the third part, the pre-SFT LLM reports the proportion of its funds invested in the stock $\{P_1^i(t)/X_1^i(t)\}_{t\in\mathcal{T}}$, based on which, we then calculate its investment decision $\{P_1^i(t)\}_{t\in\mathcal{T}}$. 

\subsection{Validation of Pre-SFT LLMs \texorpdfstring{\&}{and} the Theoretical Solution}
\label{sec: comparison}

\begin{figure*}[!t]
    \centering
    \subfigure[\texttt{GPT-3.5}]
    {
    \includegraphics[width=0.315\textwidth]{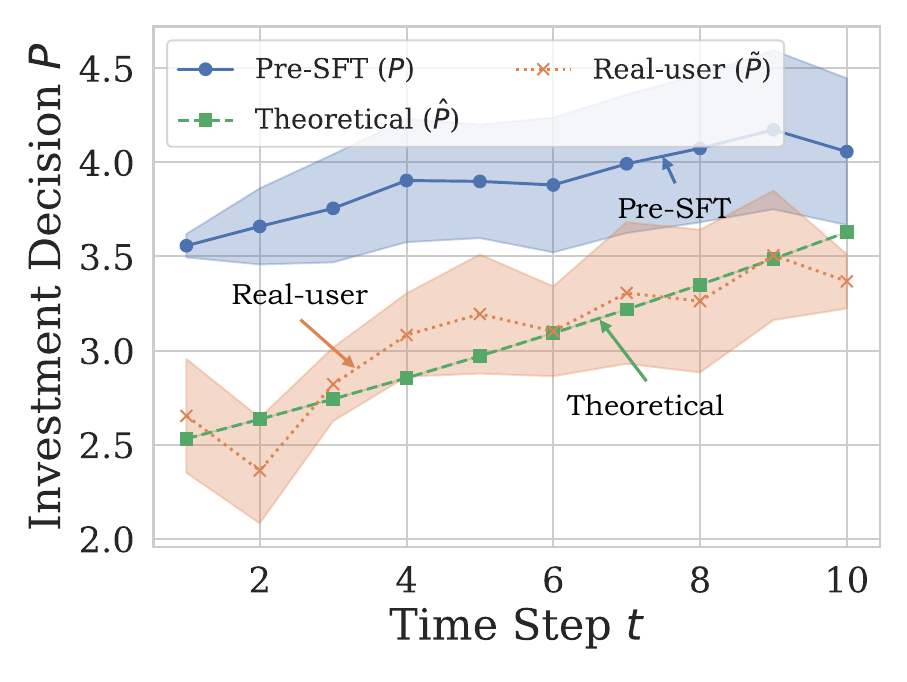}
    \label{fig: gpt_time_step}
    }
    \subfigure[\texttt{Qwen-2}]
    {
    \includegraphics[width=0.315\textwidth]{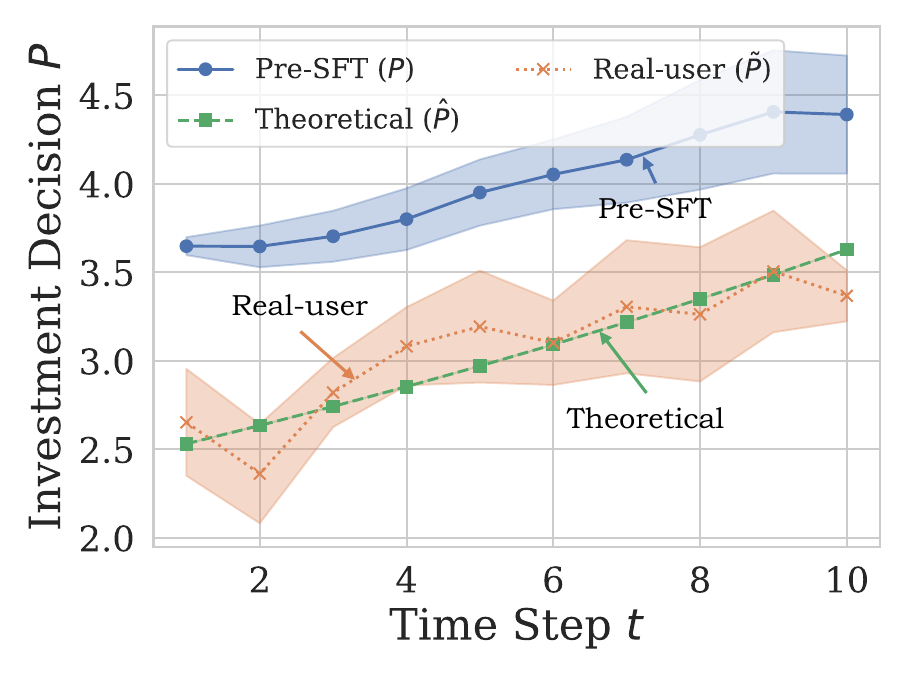}
    \label{fig: qwen_time_step}
    }
    \subfigure[\texttt{Llama-3.1}]
    {
    \includegraphics[width=0.315\textwidth]{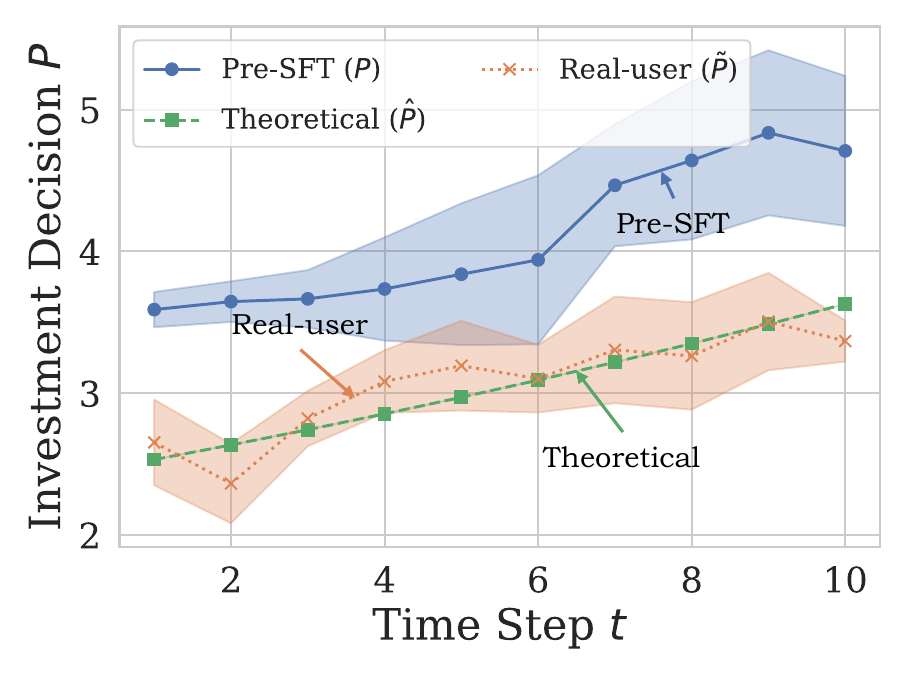}
    \label{fig: llama_time_step}
    }
        \vspace{-3mm}\caption{Comparison of real-user data ($\tilde{P}_1$), pre-SFT LLMs' decision ($P_1$), and theoretical solution ($\hat{P}_1$) on $\mathsf{P_3}$. 
    }
    \label{fig: llm_as_investor}    \vspace{-3mm}
\end{figure*}

The real-user data shows that the participants' risk aversion coefficients $\{\alpha_1^i\}_{i\in\mathcal{I}}$ and influence coefficients $\{\theta_1^i\}_{i\in\mathcal{I}}$ fall within the ranges of $\tilde{\mathcal{S}}_{\alpha_1}=[0.09, 0.38]$ and $\tilde{\mathcal{S}}_{\theta_1}=[0, 1\times10^{-7}]$, respectively. For the convenience of data processing, we discretize these two sets into $\tilde{\mathcal{S}}_{\alpha_1}=\bigcup_{m\in\mathcal{M}}\tilde{\mathcal{S}}_{\alpha_1}^m$ and $\tilde{\mathcal{S}}_{\theta_1}=\bigcup_{n\in\mathcal{N}}\tilde{\mathcal{S}}_{\theta_1}^n$, and treat values that fall within the same interval as the same value\footnote{Specifically, we set $\tilde{\mathcal{S}}_{\alpha_1}=[0.09,0.13)\cup[0.13,0.19)\cup[0.19,0.26)\cup[0.26,0.38)\cup\{0.38\}$ and $\tilde{\mathcal{S}}_{\theta_1}=\bigcup_{k=0}^9[k\times10^{-8},(k+1)\times10^{-8})\cup\{1\times10^{-7}\}$, and use the left-point value to approximate the entire interval.}. We then group the participants according to these subsets, with participants sharing the same investment attributes forming a class. Specifically, the class of participants with risk aversion coefficient $\alpha_1\in\tilde{\mathcal{S}}_{\alpha_1}^m$ and influence coefficient $\theta_1\in\tilde{\mathcal{S}}_{\theta_1}^n$ for all $m\in\mathcal{M}$ and $n\in\mathcal{N}$ is denoted as $\mathcal{I}^{mn}=\{i|\alpha_1^i\in\tilde{\mathcal{S}}_{\alpha_1}^m,\theta_1^i\in\tilde{\mathcal{S}}_{\theta_1}^n\}$ for all $m\in\mathcal{M}$ and $n\in\mathcal{N}$.

For each participant class $\mathcal{I}^{mn}$, we calculate the mean and the $95\%$ confidence interval of the real-user data, the mean and the $95\%$ confidence interval of the pre-SFT LLMs' investment decisions based on $10$ repeated trials with the same investment attribute, and the corresponding theoretical solution. Here, we take the investment attribute $\alpha_1=0.13$ and $\theta_1=7\times 10^{-8}$ as an example, and observe the same trend for other values. The experimental results are in Figure \ref{fig: llm_as_investor}. In figure legends, we omit the subscript $1$ from $P_1$ where no ambiguity arises. As shown in Figure \ref{fig: llm_as_investor}, there is a significant discrepancy between the pre-SFT LLMs' investment decisions and the real-user data, indicating that pre-SFT LLMs fail to align with real-user data in optimal investment under absolute herd behavior. We also find that the performances of pre-SFT LLMs in $\mathsf{P_1}$ and $\mathsf{P_2}$ are misaligned, as shown in Appendix \ref{sec:P1}, underscoring the necessity of supervised fine-tuning to bridge the gap between pre-SFT LLMs' decisions and real-user data. 

On the contrary, from Figure \ref{fig: llm_as_investor}, the theoretical solutions are much closer to the real-user data than pre-SFT LLMs' investment decisions. We further employ statistical methods to validate the consistency between the theoretical solutions and real-user data. For the $i$-th participant, we denote his/her real investment decision as $\{\tilde{P}_1^i(t)\}_{t\in\mathcal{T}}$, and the theoretical solution with the same investment attribute as $\{\hat{P}_1^i(t)\}_{t\in\mathcal{T}}$, respectively. We calculate the difference $d_i$ and correlation coefficient $\rho_i$ between $\{\tilde{P}_1^i(t)\}_{t\in\mathcal{T}}$ and $\{\hat{P}_1^i(t)\}_{t\in\mathcal{T}}$, and conduct $t$-tests on the means of the differences $\{d_i\}_{i\in\mathcal{I}}$ and the correlation coefficients $\{\rho_i\}_{i\in\mathcal{I}}$ \cite{shao2008mathematical}, respectively. The experimental results in Appendix \ref{sec:statistical} show that there exists significant consistency between the theoretical solution and real-user data. 

In summary, due to the significant gap between pre-SFT LLMs and real-user data, fine-tuning the LLMs with the theoretical solution is critical. As the theoretical solution closely aligns with real-user data, we can use them to construct the SFT dataset as a substitute for real-user data.

\section{Methodology: \texorpdfstring{\textbf{InvestAlign}}{InvestAlign}}
\label{sec:method}
In this section, to answer $\mathsf{Q_3}$, i.e., how we can construct the SFT dataset using the theoretical solution, and whether it performs better in fine-tuning compared to real-user data, we first introduce the method of constructing SFT datasets using the theoretical solution. Then, we theoretically prove that training LLMs on these datasets results in faster parameter convergence than using real-user data.

\subsection{Constructing SFT Dataset with Theoretical Solution}
\label{sec:sft}
The SFT dataset comprises input-output pairs for fine-tuning LLMs, which are generated based on a custom prompt template. The prompt for SFT is in Figure \ref{fig:sft_absolute} in Appendix \ref{sec:prompt}. When constructing the SFT dataset, we need to vary $\mathsf{A_1}$'s investment attribute $\alpha_1$ and $\theta_1$. Following the work in \cite{wang2024herd}, we set the values of $\alpha_1$ and $\theta_1$ in $\hat{\mathcal{S}}_{\alpha_1}=\{0.05,0.10,\ldots,0.50\}$ and $\hat{\mathcal{S}}_{\theta_1}=\{1\times10^{-8},2\times10^{-8},\ldots,1\times10^{-7}\}$. Using the same method in Section \ref{sec: data}, we set the investment attributes through two questions in natural language easy for LLMs to understand, rather than directly telling them the specific values of the parameters. For each investment attribute, we first calculate the theoretical optimal decision $\{\hat{P}_1(t)\}_{t\in\mathcal{T}}$ using (\ref{eq:theoretical}) and Algorithm \ref{alg:1}, and then calculate the investment proportion $\{\hat{P}_1(t)/X_1(t)\}_{t\in\mathcal{T}}$ using (\ref{eq:budget}). Note that there exists a random perturbation $\{W(t)\}_{t\in\mathcal{T}}$ in (\ref{eq:budget}), and we repeat $10$ trials for each investment attribute. In summary, the SFT dataset contains $10^3$ training samples.

\subsection{Analysis of Parameter Convergence Rates}
\label{sec:theoretical}
We theoretically show that fine-tuning LLMs on the training datasets constructed from theoretical solutions leads to faster parameter convergence compared to using real-user data. To ensure mathematical tractability, we make the following assumptions. First, when calculating the loss function, we only consider the values of the LLM's investment decision, theoretical solution, and real-user data. This is because the natural language parts for all three experiments are the same. Second, we assume that the sample size of the training dataset constructed from the theoretical solution and real-user data are both sufficiently large. Third, we assume that the output layer of the LLM is a Sigmoid layer, i.e., $\mathrm{Sigmoid}(\mathbf{z})=[1+\exp(-\mathbf{w}^\top \mathbf{z})]^{-1}$, where $\mathbf{w}$ is the model parameter and $\mathbf{z}$ is the output layer's input. Although the output layer of the LLM may be more complex, this simplification makes the theoretical analysis tractable. Fourth, we assume a locally convex loss landscape near the optimum, which a standard premise in optimization theory \cite{boyd2004convex}. We denote the ranges of the LLM's investment decision $P_1(t)$, theoretical solution $\hat{P}_1(t)$, and real-user data $\tilde{P}_1(t)$ as $\mathcal{P}_1(t)$, $\hat{\mathcal{P}}_1(t)$, and $\tilde{\mathcal{P}}_1(t)$, respectively. 

Given the above assumptions, in the following, we analyze the parameter convergence rate in fine-tuning. First, according to the second assumption, when fine-tuning the LLM using the training dataset constructed from the theoretical solution, we can express the cross-entropy loss function as
\begin{equation}\textstyle
    \!\!\!\!\!\hat{L}(\mathbf{w})=-\!\sum_{t\in\mathcal{T}}\!\int_{\hat{\mathcal{P}}_1(t)}\!f_{\hat{P}_1(t)}(x)\ln \!f_{P_1(t)}(x)\dif x,\!\!\!\label{eq:loss}
\end{equation}
where $f_{P_1(t)}(\cdot)$ and $f_{\hat{P}_1(t)}(\cdot)$ represent the probability density functions of $P_1(t)$ and $\hat{P}_1(t)$ in the training dataset, respectively. Similarly, we can define the cross-entropy loss function $\tilde{L}(\mathbf{w})$ when fine-tuning the LLM using the real-user data. 

Next, we derive the analytical form of $f_{\hat{P}_1(t)}(\cdot)$ and $f_{\tilde{P}_1(t)}(\cdot)$. When we construct the SFT dataset, we uniformly set the values of the risk aversion coefficient $\alpha_1$ and the influence coefficient $\theta_1$ within a rectangular region. Therefore, we assume that $\alpha_1$ and $\theta_1$ satisfy two independent uniform distributions. As shown in Appendix \ref{sec:probability}, we can prove that for all $x\in\hat{\mathcal{P}}_1(t)$, 
\begin{equation}\textstyle
    f_{\hat{P}_1(t)}(x)\approx cx^{-2},\label{equ: pdf}
\end{equation}
where $c$ is a normalization parameter. Equation (\ref{equ: pdf}) is consistent with the empirical research in the field of behavioral finance, which shows that the distribution of trading volume often exhibits a power-law characteristic \cite{iori2002microsimulation}. From (\ref{equ: pdf}), we can find that the probability distribution function $f_{\hat{P}_1(t)}(\cdot)$ is a monotonically decreasing function. Thus, we assume that $f_{P_1(t)}(\cdot)$ is also monotonically decreasing. Because the real-user data often have a bigger noise than the theoretical solution, we assume that $\tilde{P}_1(t)$ is $\hat{P}_1(t)$ plus a white noise $n(t)$ that independently and identically satisfies a uniform distribution $U(-\varepsilon,\varepsilon)$, i.e, $\tilde{P}_1(t)=\hat{P}_1(t)+n(t)$. Then, as shown in Appendix \ref{sec:convergence}, we can derive the expression of $f_{\hat{P}_1(t)}(\cdot)$. Finally, given $f_{\hat{P}_1(t)}(\cdot)$ and $f_{\tilde{P}_1(t)}(\cdot)$, we can calculate the gradient norms of the loss function and further prove that
\begin{equation}\textstyle
    \Vert\nabla\hat{L}(\mathbf{w})\Vert>\Vert\nabla\tilde{L}(\mathbf{w})\Vert.\label{eq:gradcom}
\end{equation}
That is, the gradient norm when using the training dataset constructed from the theoretical solution is higher than when using real-user data. This is because, once the parameters are given, the real-user data are noisy, while the theoretical solution is deterministic. According to \cite{chen2018survey} and the fourth assumption, a larger gradient norm leads to faster convergence due to steeper descent directions aligned with the theoretical solution's distribution in the early training phases when parameters reside in convex-like regions. Thus, from (\ref{eq:gradcom}), we can draw the conclusion that the gradient descent algorithm converges faster when using the training dataset compared to using real-user data.

\begin{figure*}[!t]
    \centering
    \subfigure[\texttt{Qwen-2}]{\includegraphics[width=.315\textwidth]{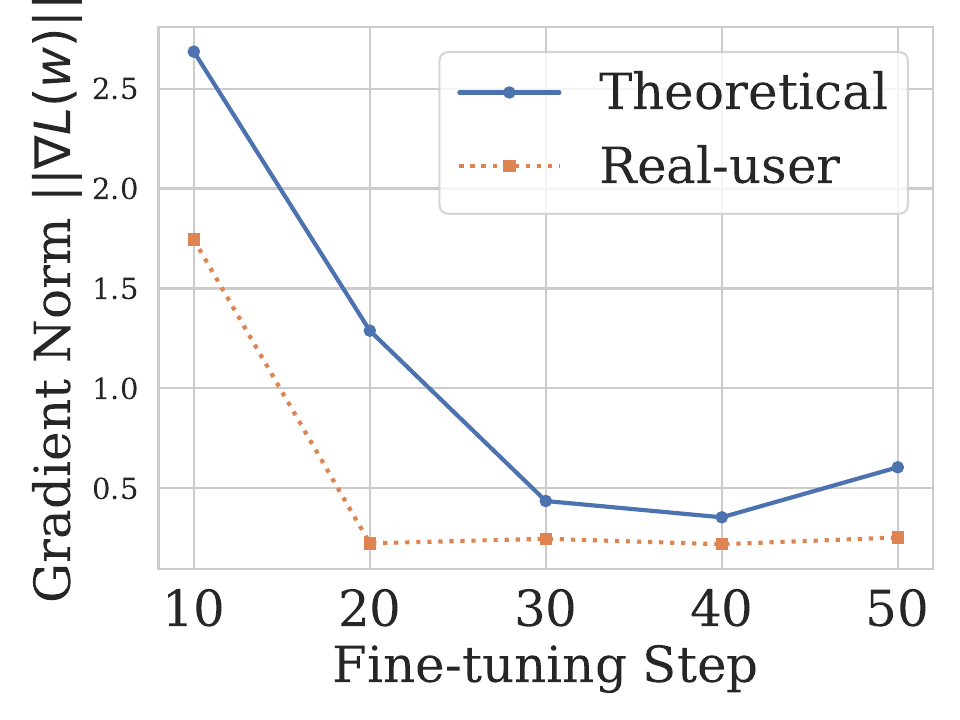}\label{fig: qwen2-grad}}
    \subfigure[\texttt{Llama-3.1}]{\includegraphics[width=.315\textwidth]{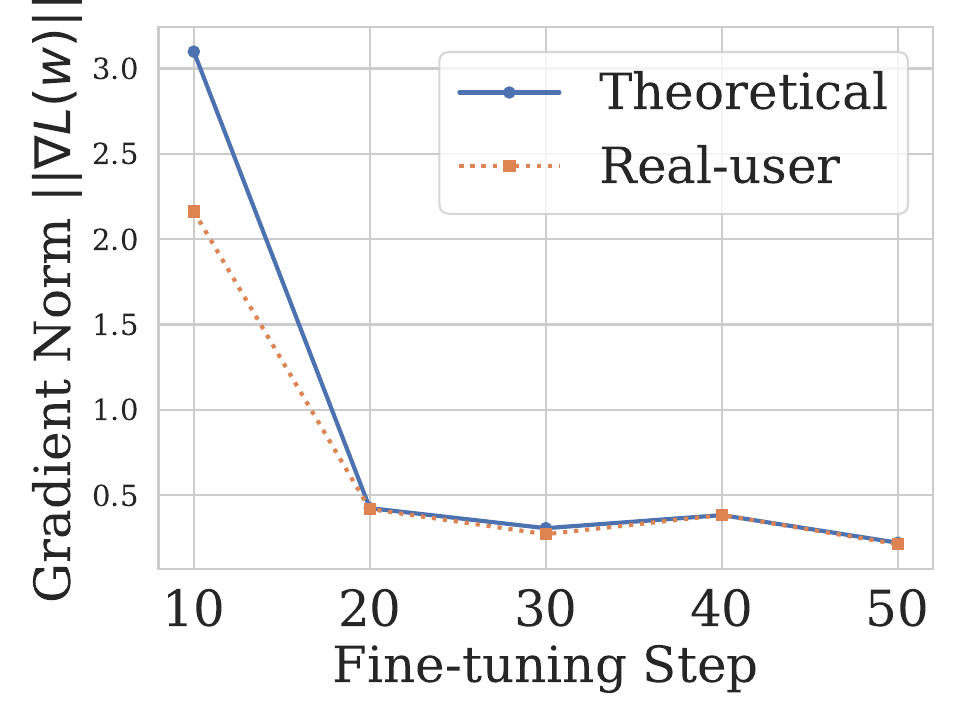}\label{fig: llama-grad}}
        \vspace{-3mm}\caption{Comparison of the gradient norms between using theoretical solution and real-user data.}
    \label{fig: grad_norm}
\end{figure*}

\begin{figure*}[!t]
    \centering
    \subfigure[\texttt{Qwen-2}]
    {\includegraphics[width=0.315\textwidth]{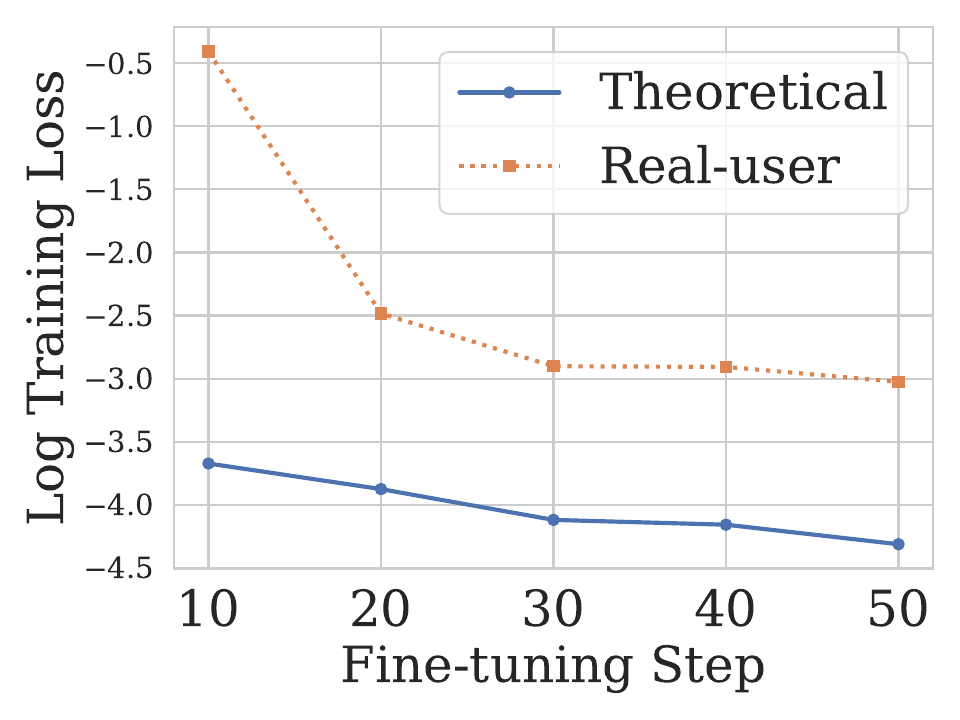}}
    \subfigure[\texttt{Llama-3.1}]{\includegraphics[width=0.315\textwidth]{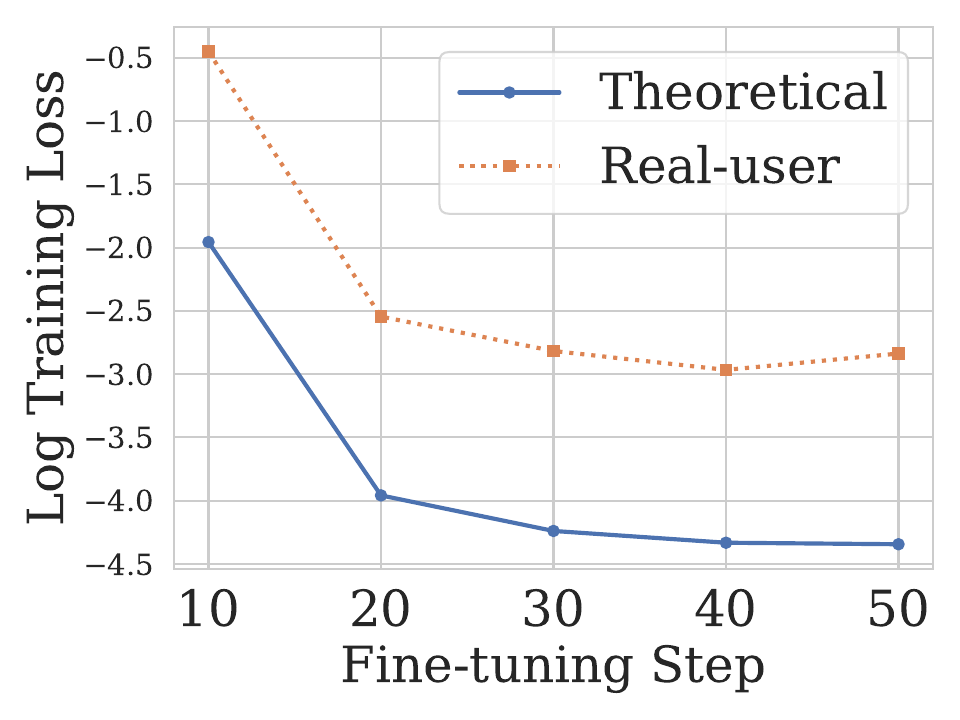}}
        \vspace{-3mm}\caption{Comparison of the logarithmic training losses between using theoretical solution and real-user data.}
    \label{fig:revision-fig1}
\end{figure*}

We conduct an experiment to validate our above analysis on open-source models including \texttt{Qwen-2} and \texttt{Llama-3.1}. We construct the SFT datasets using both the theoretical solution and real-user data, and fine-tune the LLMs with these training datasets using low-rank adaptation (LoRA) in \cite{hu2021lora}. We set the LoRA rank, alpha, and dropout rate as $4$, $32$, and $0.1$, respectively, and keep the training parameters, such as the learning rate and batch size, etc., unchanged. The experimental results of the gradient norm $\Vert\nabla L(\mathbf{w})\Vert$ are in Figure \ref{fig: grad_norm}. From Figure \ref{fig: grad_norm}, the gradient norm when using the training dataset constructed from theoretical solution is significantly higher than when using real-user data across different LLMs, validating that fine-tuning LLMs on the training datasets constructed from theoretical solution leads to faster parameter convergence than using real-user data.

We also conduct an experiment to show the relationship between training loss reductions and fine-tuning steps for models trained on theoretical and real-user data to validate our analysis of parameter convergence rates. The experimental results are in Figure \ref{fig:revision-fig1}. From Figure \ref{fig:revision-fig1}, \textbf{InvestAgent}s trained with theoretical data exhibit substantially lower logarithmic training losses and faster loss decay compared to those trained with real-user data, aligning with our gradient norm analysis.

\section{Experiments \texorpdfstring{\&}{and} Performance Validation}
\label{sec:experiment}
To answer $\mathsf{Q_4}$, we conduct experiments to verify the alignment performance of \textbf{InvestAgent}s with real-user data in the simple problem $\mathsf{P_3}$ and the original complex problems $\mathsf{P_1}$ and $\mathsf{P_2}$.

\subsection{Performance of \texorpdfstring{\textbf{InvestAgent}}{InvestAgent} in \texorpdfstring{$\mathsf{P_3}$}{P3}}
\textbf{Experimental Setup.}
\label{sec:exp_align}
To compare the alignment performance of pre-SFT LLMs and \textbf{InvestAgent}s with real-user data, we develop a Python-based simulation environment. The prompt we used is in Figure \ref{fig:prompt_absolute} in Appendix \ref{sec:prompt}. For different investment attributes, we select $\alpha_1$ from $\mathcal{S}_{\alpha_1} = \{0.09, 0.13, 0.19, 0.26, 0.38\}$ and $\theta_1$ from $\mathcal{S}_{\theta_1} = \{0, 1\times10^{-8}, \dots, 1\times10^{-7}\}$. Given $\{W(t)\}_{t\in\mathcal{T}}$ in (\ref{eq:budget}), we use $10$ random seeds for each investment attribute, producing a total of $550$ trials.

\begin{figure*}[!t]
    \centering
    \subfigure[\texttt{GPT-3.5}]
    {
    \includegraphics[width=0.315\textwidth]{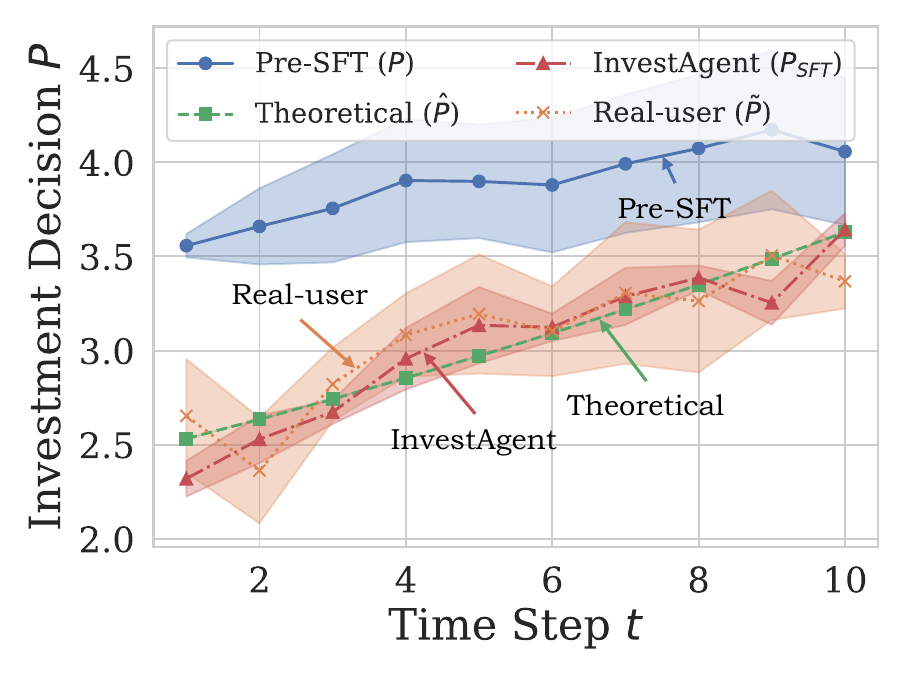}
    \label{fig: gpt}
    }
    \subfigure[\texttt{Qwen-2}]
    {
    \includegraphics[width=0.315\textwidth]{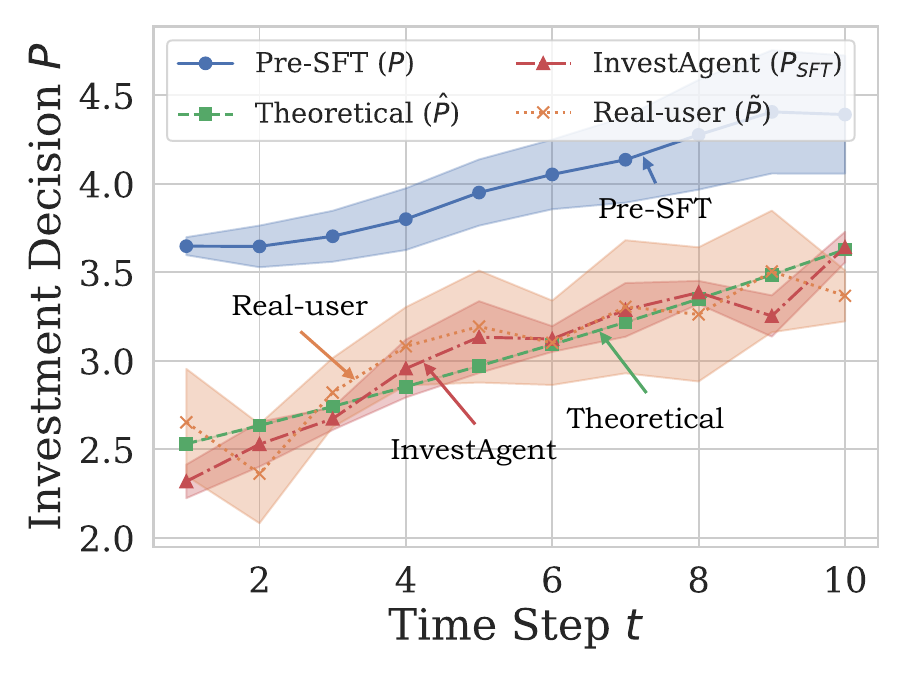}
    \label{fig: qwen2}
    }
    \subfigure[\texttt{Llama-3.1}]
    {
    \includegraphics[width=0.315\textwidth]{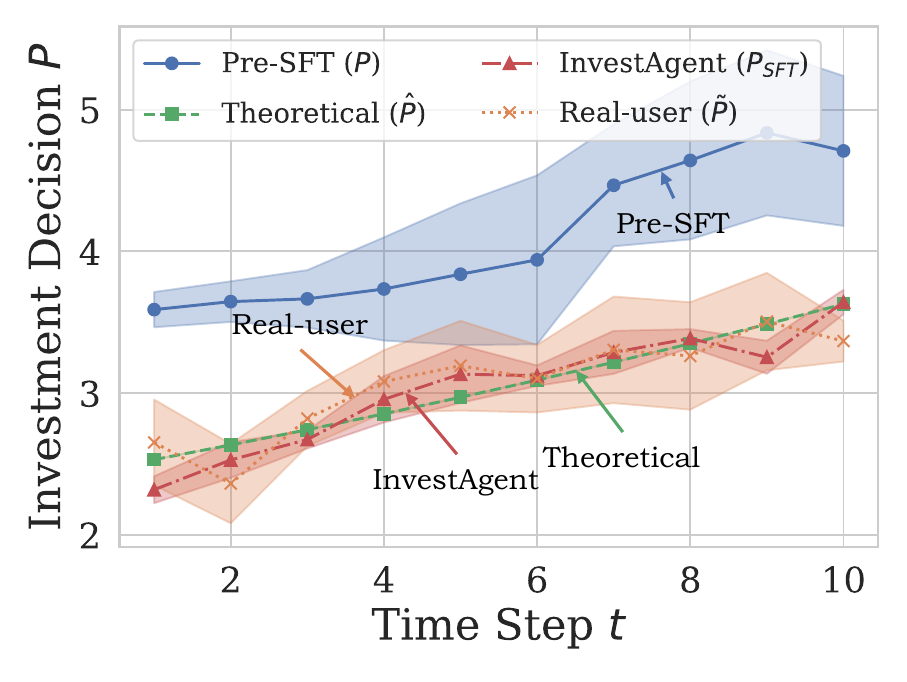}
    \label{fig: llama}
    }
        \vspace{-3mm}\caption{Comparison of real-user data ($\tilde{P}_1$), pre-SFT LLMs' investment decision ($P_1$), \textbf{InvestAgent}s' investment decision ($P_1^{\mathrm{SFT}}$), and theoretical solution ($\hat{P}_1$) on $\mathsf{P_3}$.
    }
    \label{fig: invest_decision}    \vspace{-3mm}
\end{figure*}

\noindent\textbf{Experimental Results.}
Similarly to the data-processing method in Section \ref{sec: comparison}, we plot the mean and the $95\%$ confidence interval of the real-user data, denoted by $\tilde{P}_1$, and the pre-SFT LLMs' and \textbf{InvestAgent}s' investment decisions based on $10$ repeated trials with the corresponding investment attribute, denoted by $P_1$, and $P_1^{\mathrm{SFT}}$, respectively. We also plot the theoretical solutions, denoted by $\hat{P}_1$. The experimental results are in Figure \ref{fig: invest_decision}. Here, we take the investment attribute $\alpha_1=0.13$ and $\theta_1=7\times 10^{-8}$ as an example, and we observe the same trend for other values. As shown in Figure \ref{fig: invest_decision}, \textbf{InvestAgent}s' investment decisions are significantly closer to real-user data and theoretical solutions than pre-SFT LLMs across different LLMs. 

Additionally, to quantitatively evaluate how \textbf{InvestAlign} can help pre-SFT LLMs align with real-user data in $\mathsf{P_3}$, we calculate the overall MSE between the mean of pre-SFT LLMs' decisions and real-user data, which is
\begin{align}
    &\textstyle\text{Overall}\ \text{MSE}(P_1,\tilde{P}_1)=\frac{1}{|\mathcal{M}||\mathcal{N}||\mathcal{T}|}\\
    &\textstyle\ \ \cdot\sum_{m\in\mathcal{M},n\in\mathcal{N},t\in\mathcal{T}}[P_1^{mn}(t)-\tilde{P}_1^{mn}(t)]^2,\notag
\end{align}
and the overall MSE between the mean of \textbf{InvestAgent}s' decisions and real-user data, which is
\begin{align}
    &\textstyle\text{Overall}\ \text{MSE}(P_1^{\mathrm{SFT}},\tilde{P}_1)=\frac{1}{|\mathcal{M}||\mathcal{N}||\mathcal{T}|}\\
    &\textstyle\ \ \ \cdot\sum_{m\in\mathcal{M},n\in\mathcal{N},t\in\mathcal{T}}[P_1^{\mathrm{SFT},mn}(t)-\tilde{P}_1^{mn}(t)]^2,\notag
\end{align}
where $\{\tilde{P}_{mn}(t)\}_{t\in\mathcal{T}}$ represents the mean of the real-user data in class $\mathcal{I}^{mn}$, $\{P_{mn}(t)\}_{t\in\mathcal{T}}$ and $\{P_{SFT,mn}(t)\}_{t\in\mathcal{T}}$ represents the mean of the pre-SFT LLMs' and \textbf{InvestAgent}s' investment decisions with the corresponding investment attribute, respectively. 
The experimental results are in Table \ref{tab: overall_mse}. As shown in Table \ref{tab: overall_mse}, \textbf{InvestAlign} helps reduce the overall MSEs by $45.59\%\sim61.26\%$. Note that the pre-SFT LLM's overall MSE for $\mathrm{P_3}$ is inherently lower than those for $\mathrm{P_1}$ and $\mathrm{P_2}$ due to the simplicity of $\mathrm{P_3}$. Hence, the reduction space of overall MSE is narrower for $\mathrm{P_3}$ than $\mathrm{P_1}$ and $\mathrm{P_2}$.

We also conduct an ablation study on the hyper-parameters of fine-tuning, including LoRA Rank and fine-tuning steps, as shown in Appendix \ref{sec:ablation}. We find that the overall MSE decreases as either LoRA Rank or fine-tuning steps increase.

The above results validate the effectiveness of our proposed method \textbf{InvestAlign}, i.e., fine-tuning LLMs using the SFT dataset constructed from the theoretical solution can align them better with investor decision-making under herd behavior.

\subsection{Performance of \texorpdfstring{\textbf{InvestAgent}}{InvestAgent} in \texorpdfstring{$\mathsf{P_1}$}{P1} \texorpdfstring{\&}{and} \texorpdfstring{$\mathsf{P_2}$}{P2}}
\textbf{Experimental Setup.}
This experiment shows the alignment performance of our proposed \textbf{InvestAlign}, i.e., using LLMs fine-tuned from $\mathsf{P_3}$ to solve $\mathsf{P_1}$ and $\mathsf{P_2}$. The prompts we use in $\mathsf{P_1}$ and $\mathsf{P_2}$ are in Figure \ref{fig:prompt_relative} and Figure \ref{fig:prompt_mutual} in Appendix \ref{sec:prompt}, respectively. The investment attributes are set the same as those in Section \ref{sec:exp_align}.

Also, we collect $80$ and $44$ real-user data for $\mathsf{P_1}$ and $\mathsf{P_2}$, respectively, and the participants are also primarily professionals and students in the fields of finance to reduce bias and noise. The questionnaires we use in $\mathsf{P_1}$ and $\mathsf{P_2}$ are in Figure \ref{fig:questionnaire_relative} and Figure \ref{fig:questionnaire_mutual} in Appendix \ref{sec:questionnaire}, respectively. Note that here, the experiments are performed on all three problems for open-source models \texttt{Qwen-2} and \texttt{Llama-3.1}, and for API-based model \texttt{GPT-3.5}, we take the simple problem $\mathsf{P_3}$ and the complex problems $\mathsf{P_1}$ and $\mathsf{P_2}$ as examples to show the experimental results. We will study the solving ability of \textbf{InvestAgent} on more complex optimal investment problems in our future work.

\begin{table}[!t]
    \vspace{1.8mm}
    \small
    \centering
    \setlength{\tabcolsep}{1.8mm}
    \begin{tabular}{c|c|cccc}
    \hline
    \multicolumn{2}{c|}{Overall MSE} & \textbf{Pre-SFT} & \textbf{InvestAgent} & \textbf{Reduction} \\
    \hline
    \multirow{3}{*}{$\mathsf{P_3}$} & \texttt{GPT-3.5} & 4.44 & 1.72 & -61.26\% \\
    & \texttt{Qwen-2} & 3.97 & 2.16 & -45.59\% \\
    & \texttt{Llama-3.1} & 4.08 & 1.59 & -61.03\% \\
    \hline
    \multirow{3}{*}{$\mathsf{P_1}$} & \texttt{GPT-3.5} & 14.03 & 7.46 & -46.84\% \\
    & \texttt{Qwen-2} & 17.22 & 7.46 & -56.69\% \\
    & \texttt{Llama-3.1} & 13.07 & 7.25 & -44.52\% \\
    \hline
    \multirow{2}{*}{$\mathsf{P_2}$} & \texttt{Qwen-2} & 15.66 & 6.12 & -60.92\% \\
    & \texttt{Llama-3.1} & 12.28 & 6.66 & -45.77\% \\
    \hline
    \end{tabular}
    \caption{Comparison of the overall MSE between pre-SFT LLMs' and \textbf{InvestAgent}s' investment decisions with real-user data in $\mathsf{P_1}$, $\mathsf{P_2}$, and $\mathsf{P_3}$.}
    \label{tab: overall_mse}
    \vspace{-3mm}
\end{table}

\noindent\textbf{Experimental Results.}
Using the same method in Section \ref{sec:exp_align}, we list the overall MSE between the mean of pre-SFT LLMs' investment decisions with real-user data, $\text{Overall MSE}(P_1,\tilde{P}_1)$, and the overall MSE between the mean of \textbf{InvestAgent}s' investment decisions with real-user data, $\text{Overall MSE}(P_1^{\mathrm{SFT}},\tilde{P}_1)$, for $\mathsf{P_1}$ in Table \ref{tab: overall_mse}. For $\mathsf{P_2}$, we list $\text{Overall MSE}(P_2\cup P_3,\tilde{P}_2\cup\tilde{P}_3)$ and $\text{Overall MSE}(P_2^{\mathrm{SFT}}\cup P_3^{\mathrm{SFT}},\tilde{P}_2\cup\tilde{P}_3)$ correspondingly. As shown in Table \ref{tab: overall_mse}, \textbf{InvestAlign} helps reduce the overall MSEs by $44.53\%\sim56.68\%$ and $45.77\%\sim60.92\%$ in $\mathsf{P_1}$ and $\mathsf{P_2}$, respectively. The experimental results validate the effectiveness of our proposed \textbf{InvestAlign}, and show that the \textbf{InvestAgent}s fine-tuned using the theoretical solution in a similar and simple problem can better align with human decision-making processes in a complex problem than pre-SFT LLMs. It demonstrates the potential of \textbf{InvestAlign} to solve complex optimal investment problems and align LLMs with investor decision-making processes.

In addition to the experiments mentioned above, we also: 1) supplement smaller samples of real-user data with theoretical solutions to construct a training dataset to improve robustness; 2) compare \textbf{InvestAgent}s with LLMs fine-tuned using the baseline FinGPT dataset \cite{yang2023fingpt}; and 3) validate that \textbf{InvestAgent}s can reflect economic principles in the presence of investor herd behavior. The experimental results and analyses are in Appendices \ref{sec:exp1}--\ref{sec:exp3}, respectively.

\section{Conclusion}
\label{sec:conclusion}
Studying investor decision-making processes under the herd behavior is of great significance in microeconomics and behavioral finance. LLMs can be leveraged to assist in solving complex investment problems. To fine-tune LLMs for alignment with human decision-making processes, a substantial amount of real-user data is required. However, the cost of collecting the real-user data is high, and there are concerns regarding privacy and security. To address data scarcity, we propose \textbf{InvestAlign}, a novel method that constructs training datasets using the theoretical solution of a similar and simple problem to align LLMs with investor behavior. We demonstrate that fine-tuning LLMs on these training datasets leads to faster parameter convergence compared to using real-user data. The experimental results indicate that \textbf{InvestAgent}s, fine-tuned with \textbf{InvestAlign}, achieves superior alignment performance in the original complex problem. 

\section*{Limitations}
As a preliminary work in this field, \textbf{InvestAlign} does not claim universal applicability for all complex optimal investment problems, and the theoretical solution may not fully capture all nuances of real-world investor behavior. We aim to address a specific challenge: data scarcity when training LLMs in the context of investor decision-making. In our future work, we plan to: 1) extend InvestAlign to diverse behavioral biases, e.g., overconfidence and loss aversion; 2) incorporate RLHF on \textbf{InvestAgent}s to complement SFT to assess the effectiveness of different alignment methods in investment decision-making tasks.

\section*{Ethics Statement}
In our study, participants were primarily recruited from professional and student populations within the fields of microeconomics and behavioral finance. Our participant selection strategy was designed to address the constraints of data quality requirements and experimental practicality. Note that high-quality data were required to test the validity of the theoretical model. Due to the time constraint, we could only allow each participant a limited time in the experiment. Less-experienced participants might misunderstand key terms or require excessive time to make decisions. Therefore, we decided to recruit professionals, students, and experienced investors to collect high-quality data. In real-world settings, investors typically make decisions after extended observation and careful deliberation. Therefore, recruiting experienced participants for short-term experiments maintained reasonable data representativeness. 

To ensure diversity and representativeness, we employed targeted recruitment strategies, such as reaching out through academic institutions and professional networks. Regarding compensation, participants were remunerated according to the standard rates for similar studies in the respective regions. Payment levels were carefully considered based on participants' demographic characteristics to ensure fair compensation for their time and expertise. Participants were fully informed about the study’s objectives, how their data would be used, and their rights to withdraw from the study at any time without penalty. We have thoroughly reviewed the real-user data to ensure it is free from potential discrimination, human rights violations, bias, exploitation, or any other ethical concerns. Additionally, the data does not contain any information that could identify individuals or include offensive content. The data collection protocol is approved (or determined exempt) by an ethics review board.

\bibliography{custom}

\newpage

\onecolumn

\appendix
\section{Appendix}

\subsection{Theoretical Optimal Investment Decisions of \texorpdfstring{$\mathsf{P_3}$}{P3}}
\label{app:theory}
$\mathsf{A_1}$'s optimal investment decision for $\mathsf{P_3}$ is given by
\begin{equation}
    \hat{P}_1(t)=\frac{\alpha_2\sigma^2\eta\exp[2r(T-t)]+\theta_1}{\alpha_1\sigma^2\eta\exp[2r(T-t)]+\theta_1}\cdot\frac{v}{\alpha_2\sigma^2}\exp[r(t-T)],\ t\in\mathcal{T},\label{eq:theoretical}
\end{equation}
where the parameter $\eta$ can be numerically calculated using Algorithm \ref{alg:1}, and $\mathsf{A_2}$'s optimal investment decision for $\mathsf{P_3}$ is given by
\begin{equation}
    \hat{P}_2(t)=\frac{v}{\alpha_2\sigma^2}\exp[r(t-T)],\ t\in\mathcal{T}.\label{eq:hatP2}
\end{equation}
The proof can be found in \cite{wang2024herd}.

\begin{algorithm}[!ht]
    \caption{Numerical Method of the Parameter $\eta$ in $\mathsf{P_3}$.}
	\label{alg:1}
	\KwIn{Interest rate: $r$; 
 
 \quad\quad\quad Excess return rate: $v$; 
 
 \quad\quad\quad Volatility: $\sigma$; 
 
 \quad\quad\quad Initial fund: $x_{1,0}$; 
 
 \quad\quad\quad Risk aversion coefficients: $\alpha_1$ and $\alpha_2$; 
 
 \quad\quad\quad Investment period: $T$; 
 
 \quad\quad\quad Influence coefficient: $\theta_1$; 
 
 \quad\quad\quad Numerical error: $\varepsilon$.}
	\KwOut{The parameter $\eta$.}  
	\BlankLine
    $k=0$;
    
    $\eta_0=\exp\left[-\alpha_1x_{1,0}\mathrm{e}^{rT}-\frac{v^2T}{2\sigma^2}\right]$;
    
    $\Delta \eta_0=+\infty$;
    
    $\vartheta=\frac{\theta}{\alpha_1\sigma^2}$;

	\While{$\Delta\eta_k\geqslant\varepsilon$}{$\eta_{k+1}=\eta_0\exp\int_\mathcal{T}\frac{\vartheta^2v^2\left(\alpha_1/\alpha_2-1\right)^2\dif t}{2\sigma^2\left\{\eta_k\exp[2r(T-t)]+\vartheta\right\}^2}$;
 
$\Delta\eta_{k+1}=\left|\eta_{k+1}-\eta_k\right|$;
    
    $k\leftarrow k+1$;}
	$\eta\approx\eta_k$.
\end{algorithm}

\subsection{Parameter Settings}
\label{sec:para}
Following the work in \cite{wang2024herd}, we set the default parameter values in this work as:

\noindent $\bullet$ The interest rate of the deposit: $r=0.04$.

\noindent $\bullet$ The excess return rate of the stock: $v=0.03$. 

\noindent $\bullet$ The volatility of the stock: $\sigma=0.17$. 

\noindent $\bullet$ The investment period: $T=10$. 

\noindent $\bullet$ $\mathsf{A_1}$'s initial fund: $x_{1,0}=10$.

Additionally, for the optimal investment problems under absolute herd behavior with unilateral influence $\mathsf{P_1}$ and $\mathsf{P_3}$, we further set:

\noindent $\bullet$ $\mathsf{A_2}$'s risk aversion coefficient: $\alpha_2=0.2$.

\subsection{Calculation Methods of Investment Attributes}
\label{sec:attribute}
From the work in \cite{pratt1978risk}, the risk aversion coefficient $\alpha_1$ reflects the participant's preference between risky and risk-free options. If the participant is indifferent between the following two options: 1) receiving $w_1$ with probability $p_i$, and receiving nothing with probability $1 - p_i$, or 2) receiving $w_2$, his/her risk aversion coefficient $\alpha_1^i$ can be determined by
\begin{equation}
    p_i = \frac{\exp(-\alpha_1^i w_2) - 1}{\exp(-\alpha_1^i w_1) - 1}.
\end{equation}
We ask the participant to provide his/her response for $p_i$, from which we calculate his/her risk aversion coefficient $\alpha_1^i$. 

From the work in \cite{wang2024herd}, the influence coefficient $\theta_1$ quantifies the level of herd behavior. In the third part of the questionnaire, we ask participants: ``On a scale $[0,10]$, how much do you rely on the investment assistant when making decisions, where $10$ represents the highest reliance level and $0$ the lowest?'' From the work in \cite{wang2024herd}, the influence coefficient $\theta_1^i$ typically falls within the range $[0, 1\times10^{-7}]$. Thus, we set the participant's influence coefficient as $\theta_1^i= k_i\times10^{-8}$, where $k_i$ is the response.

\subsection{Experimental Results of \texorpdfstring{\texttt{GLM-4-9B-CHAT}}{GLM-4-9B-CHAT}}
\label{app:glm}
\begin{figure*}[!ht]
    \centering
    \subfigure[Figure \ref{fig: llm_as_investor}]
    {
    \includegraphics[width=0.35\textwidth]{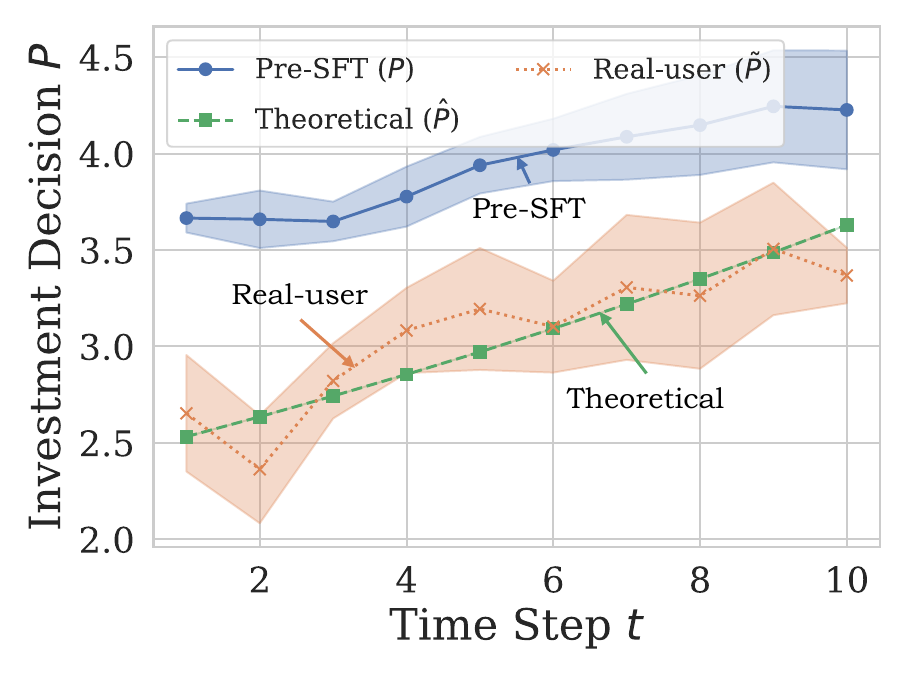}
    }
    \subfigure[Figure \ref{fig: grad_norm}]
    {
    \includegraphics[width=0.35\textwidth]{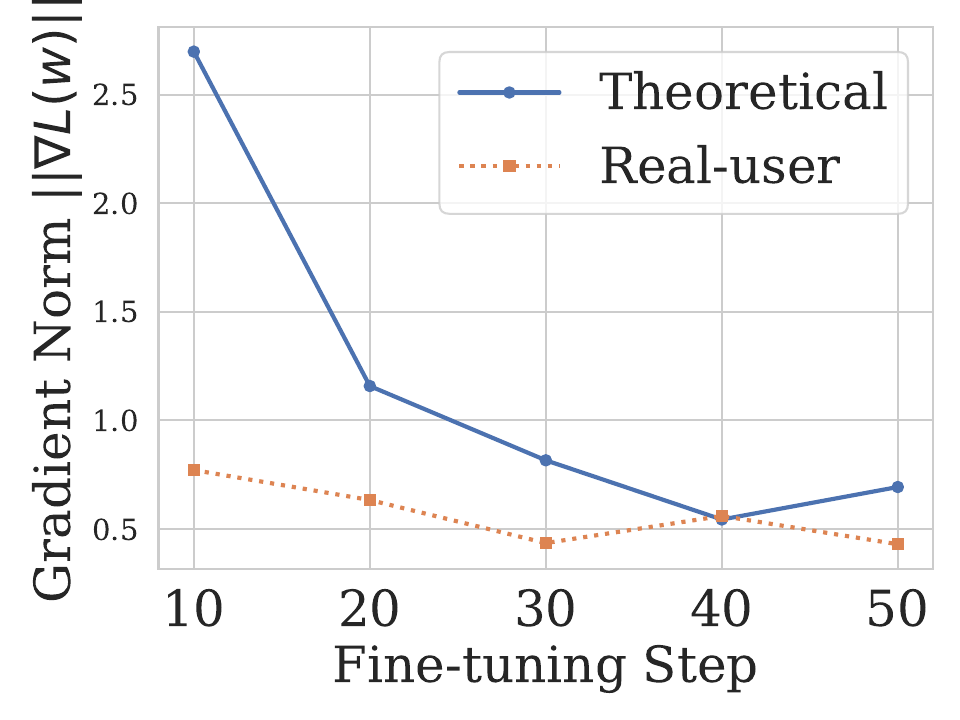}\label{fig: glm-grad}
    }
    \hfill
    \subfigure[Figure \ref{fig:revision-fig1}]
    {
    \includegraphics[width=0.35\textwidth]{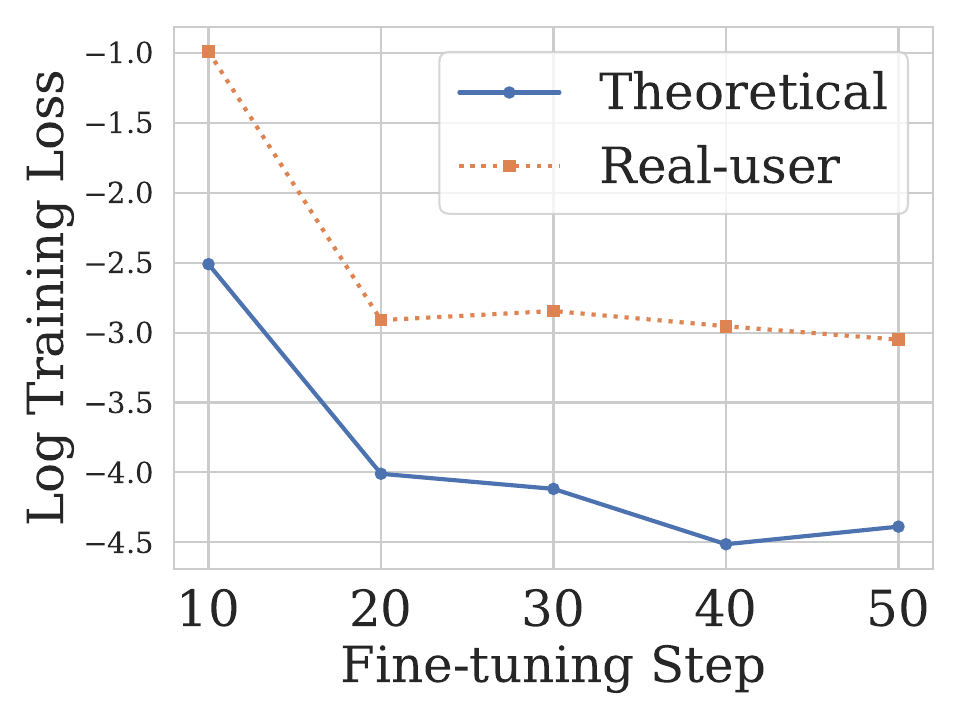}
    }
    \subfigure[Figure \ref{fig: invest_decision}]
    {
    \includegraphics[width=0.35\textwidth]{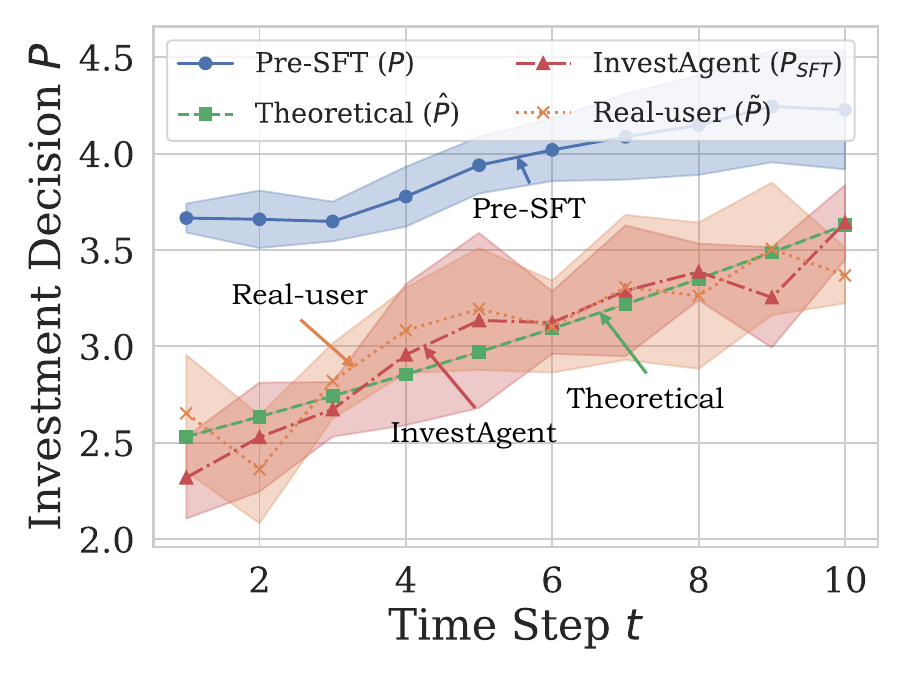}
    }
    \caption{The experimental results of \texttt{GLM-4} corresponding to Figures \ref{fig: llm_as_investor}--\ref{fig: invest_decision}.}
    \label{fig: glm}
\end{figure*}

In this work, we also conduct experiments on \texttt{GLM-4} \cite{glm2024chatglm} for better comparison. The experimental results of \texttt{GLM-4} corresponding to Figures \ref{fig: llm_as_investor}--\ref{fig: invest_decision} are in Figure \ref{fig: glm}. 

The experimental results of \texttt{GLM-4} in Figure \ref{fig: glm} are similar to those of \texttt{Qwen-2} and \texttt{Llama-3.1}, and we omit the analysis here.

\subsection{Comparison of Real-User Data and Pre-SFT LLMs' Investment Decision on \texorpdfstring{$\mathsf{P_1}$}{P1} and \texorpdfstring{$\mathsf{P_1}$}{P2}}
\label{sec:P1}
\begin{figure*}[!ht]
    \centering
    \subfigure[\texttt{GPT-3.5}]
    {
    \includegraphics[width=0.35\textwidth]{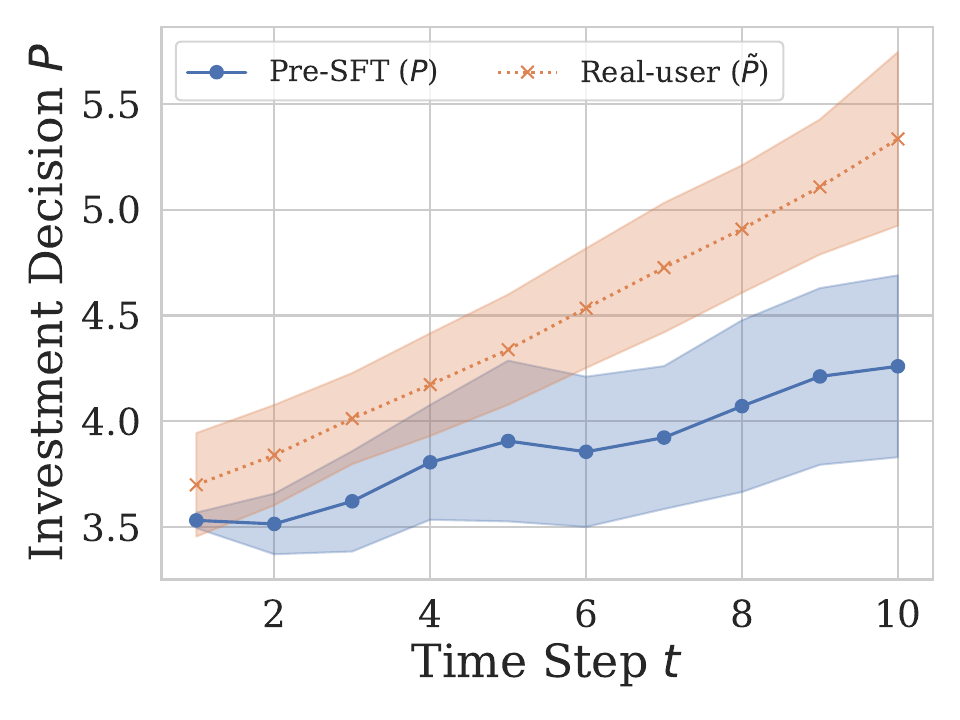}
    \label{fig: gpt_time_step_relative}
    }
    \subfigure[\texttt{GLM-4}]
    {
    \includegraphics[width=0.35\textwidth]{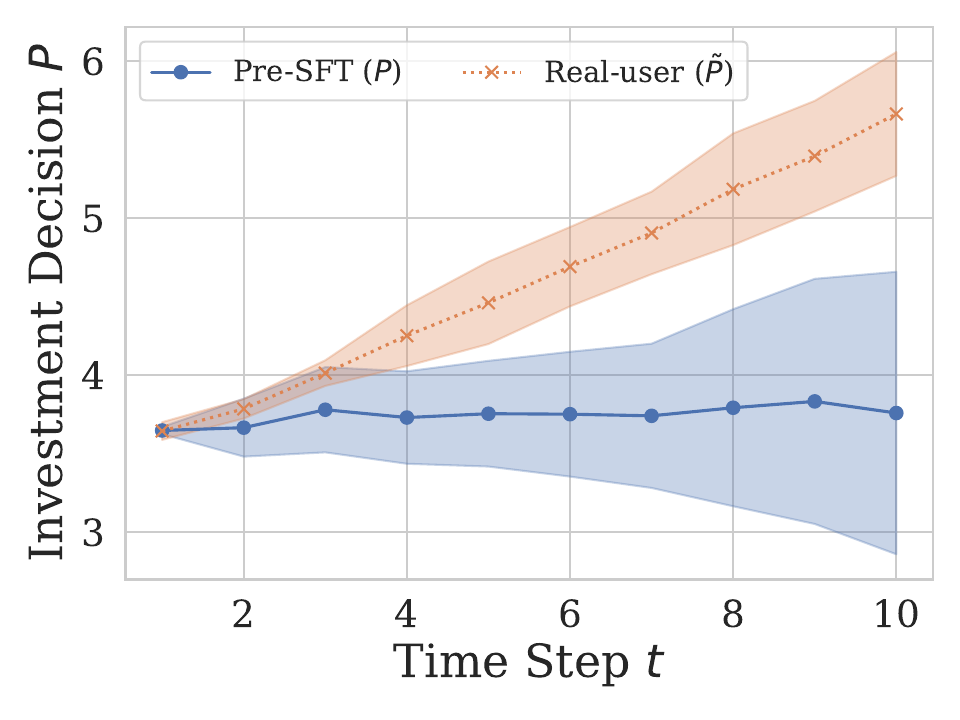}
    \label{fig: glm_time_step_relative}
    }
    \\
    \subfigure[\texttt{Qwen-2}]
    {
    \includegraphics[width=0.35\textwidth]{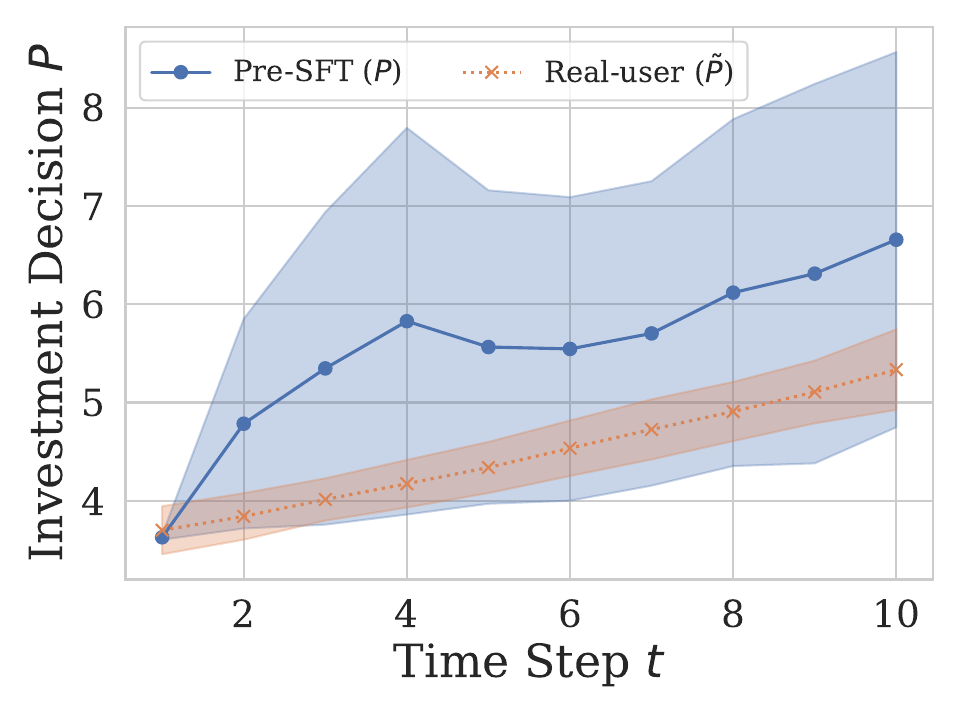}
    \label{fig: qwen_time_step_relative}
    }
    \subfigure[\texttt{Llama-3.1}]
    {
    \includegraphics[width=0.35\textwidth]{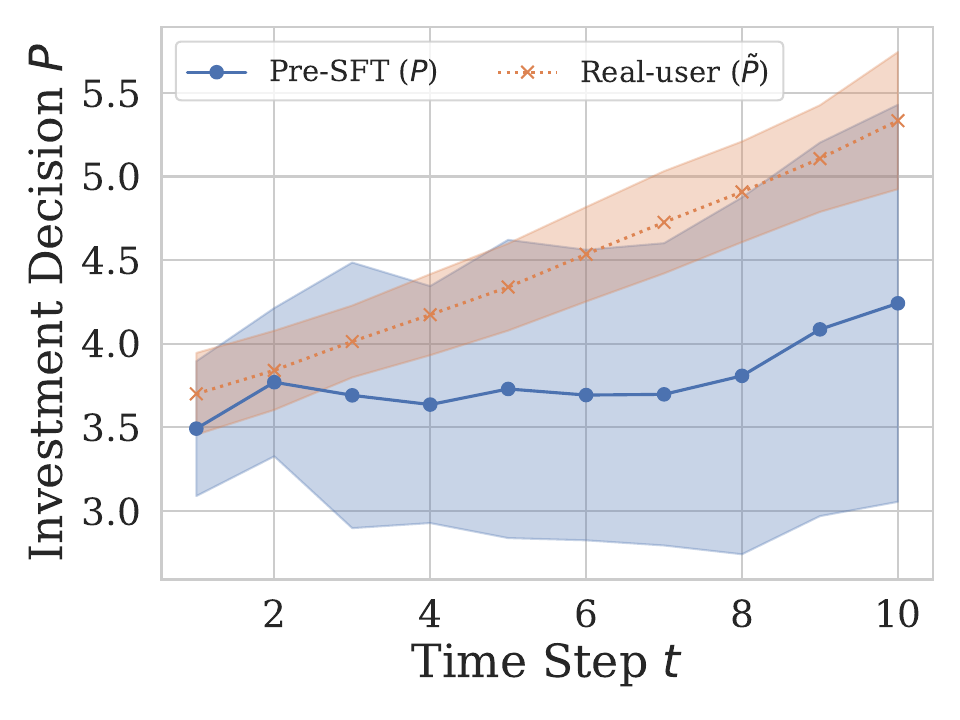}
    \label{fig: llama_time_step_relative}
    }
    \caption{Comparison of real-user data ($\tilde{P}_1$) and pre-SFT LLMs' investment decision ($P_1$) on $\mathsf{P_1}$.
    }
    \label{fig: llm_as_investor_relative}
\end{figure*}

From Figure \ref{fig: llm_as_investor_relative}, we can find that the performances of pre-SFT LLMs in $\mathsf{P_1}$ are misaligned with real-user data.
Note that when $\mathsf{A_2}$'s influence coefficient $\theta_2=0$, the complex problem $\mathsf{P_2}$ degenerates into the simple problem $\mathsf{P_3}$. Therefore, from Figure \ref{fig: llm_as_investor}, we can also draw the conclusion that the performances of pre-SFT LLMs in $\mathsf{P_2}$ are misaligned with real-user data.

\newpage

\subsection{Statistical Tests' Results of the Differences and Correlation Coefficients}
\label{sec:statistical}
We define the difference $d_i$ and correlation coefficient $\rho_i$ between $\{\tilde{P}_1^i(t)\}_{t\in\mathcal{T}}$ and $\{\hat{P}_1^i(t)\}_{t\in\mathcal{T}}$ as
\begin{equation}
    d_i=\sum_{t\in\mathcal{T}}[\tilde{P}_1^i(t)-\hat{P}_1^i(t)],
\end{equation}
and 
\begin{equation}
    \rho_i=\frac{\sum_{t\in\mathcal{T}}[\tilde{P}_1^i(t)-\bar{\tilde{P}}_1^i][\hat{P}_1^i(t)-\bar{\hat{P}}_1^i]}{\sqrt{\sum_{t\in\mathcal{T}}[\tilde{P}_1^i(t)-\bar{\tilde{P}}_1^i]^2\sum_{t\in\mathcal{T}}[\hat{P}_1^i(t)-\bar{\hat{P}}_1^i]^2}},
\end{equation}
where $\bar{\tilde{P}}_1^i=\frac{1}{T}\sum_{t\in\mathcal{T}}\tilde{P}_1^i(t)$ and $\bar{\hat{P}}_1^i=\frac{1}{T}\sum_{t\in\mathcal{T}}\hat{P}_1^i(t)$, respectively.

For the differences $\{d_i\}_{i\in\mathcal{I}}$, the results show that their mean does not significantly deviate from $0$ at the $1\%$ significance level, with a $t\text{-statistic}=-1.075$. For the correlation coefficients $\{\rho_i\}_{i\in\mathcal{I}}$, the results show that their mean does not significantly deviate from $0.85$ at the $1\%$ significance level, with a $t\text{-statistic}=-0.843$. Since a mean difference close to $0$ indicates minimal discrepancy and a correlation coefficient close to $0.85$ reflects a strong positive relationship, we show that there exists significant consistency between the theoretical solution and real-user data. 

\subsection{Probability Distribution Function of \texorpdfstring{$\mathsf{A_1}$}{A1}'s Optimal Investment Decision of \texorpdfstring{$\mathsf{P_3}$}{P3}}
\label{sec:probability}
We assume the parameters $\alpha_1$ and $\theta_1$ satisfy two uniform distributions, denoted by $U(\underline{\alpha},\overline{\alpha})$ and $U(\underline{\theta},\overline{\theta})$, respectively. Therefore, their probability distribution functions are
\begin{equation}
    f_{\alpha_1}(x)=\frac{1}{\overline{\alpha}-\underline{\alpha}},\ x\in[\underline{\alpha},\overline{\alpha}], \ \text{and}\ f_{\theta_1}(x)=\frac{1}{\overline{\theta}-\underline{\theta}},\ x\in[\underline{\theta},\overline{\theta}].
\end{equation}
From (\ref{eq:theoretical}), using the convolution formula \cite{renyi2007probability}, we have
\begin{align}
    f_{\hat{P}_1(t)}(x)&=\frac{1}{\overline{\theta}-\underline{\theta}}\int_{\underline{\theta}}^{\overline{\theta}}f_{\alpha_1}\left(\frac{1}{\sigma^2\eta\exp[2r(T-t)]}\right.\notag\\
    &\left.\cdot\left\{\frac{\alpha_2\sigma^2\eta\exp[2r(T-t)]+y}{x}\cdot\frac{v}{\alpha_2\sigma^2}\exp[r(t-T)]-y\right\}\right)\notag\\
    &\cdot\frac{\alpha_2\sigma^2\eta\exp[2r(T-t)]+y}{\sigma^2\eta\exp[2r(T-t)]x^2}\cdot\frac{v}{\alpha_2\sigma^2}\exp[r(t-T)]\dif y.\label{equ: fhatP}
\end{align}
Here, following the work in \cite{wang2024optimal}, we assume that $\eta$ approximately remains constant when $\alpha_1$ and $\theta_1$ change slightly. Because $\hat{P}_1(t)\in\hat{\mathcal{P}}_1(t)$, we can rewrite (\ref{equ: fhatP}) as
\begin{equation}
    f_{\hat{P}_1(t)}(x)\approx\frac{\min[\hat{\mathcal{P}}_1(t)]\cdot\max[\hat{\mathcal{P}}_1(t)]}{\max[\hat{\mathcal{P}}_1(t)]-\min[\hat{\mathcal{P}}_1(t)]}\cdot\frac{1}{x^2},\ x\in\hat{\mathcal{P}}_1(t).
\end{equation}
That is, $\mathsf{A_1}$'s theoretical optimal decision $\hat{P}_1(t)$ approximately satisfies a Pareto distribution. To simplify the notation, we denote the normalization parameter as
\begin{equation}
    c=\frac{\min[\hat{\mathcal{P}}_1(t)]\cdot\max[\hat{\mathcal{P}}_1(t)]}{\max[\hat{\mathcal{P}}_1(t)]-\min[\hat{\mathcal{P}}_1(t)]}.
\end{equation}

\subsection{Gradient Norms of the Loss Function}
\label{sec:convergence}
First, we derive the expression of $f_{\tilde{P}_1(t)}(\cdot)$ from (\ref{equ: pdf}). Using the convolution formula in \cite{renyi2007probability}, we can obtain
\begin{align}
    f_{\tilde{P}_1(t)}(x)&\approx\frac{1}{2\varepsilon}\int_{-\varepsilon}^{\varepsilon}f_{\hat{P}_1(t)}(x-y)\dif y\notag\\
    &=\left\{\begin{aligned}
    &\frac{c}{2\varepsilon}\int_{\min[\hat{\mathcal{P}}_1(t)]-x}^{\varepsilon}\frac{1}{(x-y)^2}\dif y, &x\in[\min[\hat{\mathcal{P}}_1(t)]-\varepsilon,\min[\hat{\mathcal{P}}_1(t)]+\varepsilon)\\
    &\frac{c}{2\varepsilon}\int_{-\varepsilon}^{\varepsilon}\frac{1}{(x-y)^2}\dif y, &x\in[\min[\hat{\mathcal{P}}_1(t)]+\varepsilon,\max[\hat{\mathcal{P}}_1(t)]-\varepsilon)\\
    &\frac{c}{2\varepsilon}\int_{-\varepsilon}^{\max[\hat{\mathcal{P}}_1(t)]-x}\frac{1}{(x-y)^2}\dif y, &x\in[\max[\hat{\mathcal{P}}_1(t)]-\varepsilon,\max[\hat{\mathcal{P}}_1(t)]+\varepsilon]\\
    \end{aligned}\right.\notag\\
    &=\frac{c}{2\varepsilon}\left(\frac{1}{\max\{\min[\hat{\mathcal{P}}_1(t)],x-\varepsilon\}}-\frac{1}{\min\{\max[\hat{\mathcal{P}}_1(t)],x+\varepsilon\}}\right),\ x\in\tilde{\mathcal{P}}(t).\label{eq:pdftilde}
\end{align}
Next, we calculate the gradient norms. We have
\begin{align}
    \nabla\hat{L}(\mathbf{w})&=-\sum_{t\in\mathcal{T}}\int_{\hat{\mathcal{P}}_1(t)}f_{\hat{P}_1(t)}(x)\nabla\ln f_{P_1(t)}(x)\dif x\notag\\
    &=-\sum_{t\in\mathcal{T}}\int_{\hat{\mathcal{P}}_1(t)}\frac{f_{\hat{P}_1(t)}(x)}{f_{P_1(t)}(x)}\nabla\mathrm{Sigmoid}(\mathbf{z})\dif x\notag\\
    &=-\mathbf{z}\sum_{t\in\mathcal{T}}\int_{\hat{\mathcal{P}}_1(t)}f_{\hat{P}_1(t)}(x)[1-f_{P_1(t)}(x)]\dif x\notag\\
    &=-\mathbf{z}\sum_{t\in\mathcal{T}}\left[\int_{\hat{\mathcal{P}}_1(t)}f_{\hat{P}_1(t)}(x)\dif x-\int_{\hat{\mathcal{P}}_1(t)}f_{\hat{P}_1(t)}(x)f_{P_1(t)}(x)\dif x\right]\notag\\
    &=-\mathbf{z}\sum_{t\in\mathcal{T}}\left[1-\int_{\hat{\mathcal{P}}_1(t)}f_{\hat{P}_1(t)}(x)f_{P_1(t)}(x)\dif x\right].\label{eq: pdf'}
\end{align}
Therefore, the gradient norm is
\begin{equation}
    \Vert\nabla\hat{L}(\mathbf{w})\Vert=\Vert\mathbf{z}\Vert\sum_{t\in\mathcal{T}}\left[1-\int_{\hat{\mathcal{P}}_1(t)}f_{\hat{P}_1(t)}(x)f_{P_1(t)}(x)\dif x\right].\label{eq: grad}
\end{equation}
Using the same method as above, we can obtain
\begin{equation}
    \Vert\nabla\tilde{L}(\mathbf{w})\Vert=\Vert\mathbf{z}\Vert\sum_{t\in\mathcal{T}}\left[1-\int_{\tilde{\mathcal{P}}_1(t)}f_{\tilde{P}_1(t)}(x)f_{P_1(t)}(x)\dif x\right].\label{eq: grad'}
\end{equation}
Then, we compare the two gradient norms $\Vert\nabla\hat{L}(\mathbf{w})\Vert$ and $\Vert\nabla\tilde{L}(\mathbf{w})\Vert$. From (\ref{eq: grad}) and (\ref{eq: grad'}), we only need to compare the following two integrals: 
\begin{equation}
    \hat{I}=\int_{\hat{\mathcal{P}}_1(t)}f_{\hat{P}_1(t)}(x)f_{P_1(t)}(x)\dif x,
\end{equation}
and
\begin{equation}
    \tilde{I}=\int_{\tilde{\mathcal{P}}_1(t)}f_{\tilde{P}_1(t)}(x)f_{P_1(t)}(x)\dif x.
\end{equation}
Because the investment decisions of the pre-SFT LLM can be arbitrary due to the randomness of model parameters, we have
\begin{equation}
    \hat{\mathcal{P}}_1(t)\subset\tilde{\mathcal{P}}_1(t)\subset\mathcal{P}_1(t).
\end{equation}
Because $f_{P_1(t)}(\cdot)$ is monotonically decreasing, from (\ref{equ: pdf}) and (\ref{eq: pdf'}), we can prove that
\begin{equation}
    \hat{I}<\tilde{I}<1,
\end{equation}
and thus we have 
\begin{equation}
    \Vert\nabla\hat{L}(\mathbf{w})\Vert>\Vert\nabla\tilde{L}(\mathbf{w})\Vert.
\end{equation}

\subsection{The Ablation Study on the Hyper-parameters of SFT}
\label{sec:ablation}
We conduct an ablation study on the hyper-parameters of fine-tuning, including LoRA Rank and fine-tuning steps. Here, we take \texttt{Qwen-2} and \texttt{Llama-3.1} in solving $\mathsf{P}_3$ as examples. The experimental results are in Tables \ref{tab:ab-rank} and Table \ref{tab:ab-step}. It can be seen that our \textbf{InvestAlign} consistently enhances the agreement between the \textbf{InvestAgent} and real-user data across various hyperparameters. Furthermore, the overall MSE decreases as the strength of fine-tuning increases, either through a larger LoRA Rank or more fine-tuning steps, underscoring the effectiveness of \textbf{InvestAlign}. We hypothesize that full-parameter fine-tuning could yield even better results if computational resources permit, which we plan to explore in future studies.

\begin{table*}[!ht]
    \centering
    \setlength{\tabcolsep}{12.5mm}
    \begin{tabular}{c|ccc}
    \hline
    \multirow{2}{*}{Overall MSE} & \multicolumn{3}{c}{\texttt{Qwen-2}} \\
    \cline{2-4}
    & $R=4$ & $R=8$ & $R=16$ \\
    \hline
    \textbf{Pre-SFT} & 3.97 & 3.97 & 3.97 \\
    \textbf{InvestAgent} & 3.09 & 2.16 & 1.35 \\
    \textbf{Reduction} & -22.17\% & -45.60\% & -65.99\% \\
    \hline
    \hline
    \multirow{2}{*}{Overall MSE} & \multicolumn{3}{c}{\texttt{Llama-3.1}} \\
    \cline{2-4}
    & $R=4$ & $R=8$ & $R=16$ \\
    \hline
    \textbf{Pre-SFT} & 4.08 & 4.08 & 4.08 \\
    \textbf{InvestAgent} & 2.40 & 1.59 & 1.36 \\
    \textbf{Reduction} & -41.18\% & -61.03\% & -66.67\% \\
    \hline
    \end{tabular}
    \caption{Ablation study on the LoRA rank ($R$) using \texttt{Qwen-2} and \texttt{Llama-3.1}.}
    \label{tab:ab-rank}
\end{table*}

\begin{table*}[!ht]
    \centering
    \setlength{\tabcolsep}{6mm}
    \begin{tabular}{c|ccccc}
    \hline
    \multirow{2}{*}{Overall MSE} & \multicolumn{5}{c}{\texttt{Qwen-2}} \\
    \cline{2-6}
    & $S=50$ & $S=100$ & $S=150$ & $S=200$ & $S=250$ \\
    \hline
    \textbf{Pre-SFT} & 3.97 & 3.97 & 3.97 & 3.97 & 3.97 \\
    \textbf{InvestAgent} & 2.83 & 3.01 & 2.67 & 2.43 & 2.16 \\
    \textbf{Reduction} & -28.71\% & -24.18\% & -32.75\% & -38.79\% & -45.59\% \\
    \hline
    \hline
    \multirow{2}{*}{Overall MSE} & \multicolumn{5}{c}{\texttt{Llama-3.1}} \\
    \cline{2-6}
    & $S=50$ & $S=100$ & $S=150$ & $S=200$ & $S=250$ \\
    \hline
    \textbf{Pre-SFT} & 4.08 & 4.08 & 4.08 & 4.08 & 4.08 \\
    \textbf{InvestAgent} & 3.17 & 2.72 & 2.92 & 1.97 & 1.59 \\
    \textbf{Reduction} & -22.30\% & -33.33\% & -28.43\% & -51.72\% & -61.03\% \\
    \hline
    \end{tabular}
    \caption{Ablation study on the fine-tuning step ($S$) using \texttt{Qwen-2} and \texttt{Llama-3.1}.}
    \label{tab:ab-step}
\end{table*}

\newpage

\subsection{The Experimental Results of Supplementing Smaller Samples of Real-user Data with Theoretical Solutions}
\label{sec:exp1}

Here, we take the complex problem $\mathsf{P_1}$ and the simple problem $\mathsf{P_3}$ as examples. We conduct the experiments using the dataset of theoretical data and smaller samples of real-user data. The experimental results of $\mathsf{P_1}$ and $\mathsf{P_3}$ are in Table \ref{tab:exp12}.

\begin{table*}[!ht]
    \centering
    \setlength{\tabcolsep}{16mm}
    \begin{tabular}{c|cc}
    \hline
    Overall MSE & \texttt{Qwen-2} & \texttt{Llama-3.1} \\
    \hline
    \multicolumn{3}{c}{{$\mathsf{P_3}$: Absolute herd behavior with unilateral influence (simple problem)}} \\
    \hline
    \textbf{Pre-SFT LLM} & 3.97 & 4.08 \\
    \textbf{Mix-SFT LLM (1:10)} & 2.85 & 3.17 \\
    \textbf{Mix-SFT LLM (1:1)} & 2.38 & 1.76 \\
    \textbf{Mix-SFT LLM (10:1)} & \textbf{2.03} & 1.64 \\
    \textbf{InvestAgent} & 2.16 & \textbf{1.59} \\
    \hline
    \multicolumn{3}{c}{{$\mathsf{P_1}$: Relative herd behavior with unilateral influence (original complex problem)}} \\
    \hline
    \textbf{Pre-SFT LLM} & 17.22 & 13.07 \\
    \textbf{Mix-SFT LLM (1:10)} & 11.33 & 10.68 \\
    \textbf{Mix-SFT LLM (1:1)} & 9.65 & 8.98 \\
    \textbf{Mix-SFT LLM (10:1)} & \textbf{7.32} & \textbf{7.06} \\
    \textbf{InvestAgent} & 7.46 & 7.25 \\
    \hline
    \end{tabular}
    \caption{Comparison of the overall MSE between pre-SFT LLMs', mix-SFT LLMs', and \textbf{InvestAgent}s' investment decisions with real-user data. ``Mix-SFT LLM ($m:n$)'' means that LLM was fine-tuned on a training dataset where the ratio of theoretical data to real-user data is $m/n$.}
    \label{tab:exp12}
\end{table*}

From Table \ref{tab:exp12}, it can be observed that supplementing a portion of real-user data slightly improved the model's performance on average, i.e., \textbf{InvestAgent}s align more with real-user data, indicating that this approach can enhance the model's robustness to some extent. Notably, as the proportion of real-user data in the entire SFT dataset gradually increases, the robustness may improve, but the parameter convergence rate decreases. We have provided both theoretical and experimental evidence for this in Section \ref{sec:theoretical}.

\newpage

\subsection{The Experimental Results of Compare \textbf{InvestAgent}s with LLMs Fine-tuned Using the Baseline FinGPT Dataset}
\label{sec:exp2}

Here, we take the complex problem $\mathsf{P_1}$ and the simple problem $\mathsf{P_3}$ as examples. We conduct the experiments using the FinGPT datasets \cite{yang2023fingpt}, including \texttt{FinGPT-FinEval} and \texttt{FinGPT-ConvFinQA}, to fine-tune LLMs, and compare them with our proposed \textbf{InvestAgent}s. The experimental results of $\mathsf{P_1}$, $\mathsf{P_2}$, and $\mathsf{P_3}$ are in Table \ref{tab:exp22}.

\begin{table*}[!ht]
    \centering
    \setlength{\tabcolsep}{15.4mm}
    \begin{tabular}{c|cc}
    \hline
    Overall MSE & \texttt{Qwen-2} & \texttt{Llama-3.1} \\
    \hline
    \multicolumn{3}{c}{{$\mathsf{P_3}$: Absolute herd behavior with unilateral influence (simple problem)}} \\
    \hline
    \textbf{Pre-SFT LLM} & 3.97 & 4.08 \\
    \textbf{FinEval-SFT LLM} & 3.35 & 3.28 \\
    \textbf{ConvFinQA-SFT LLM} & 2.77 & 1.96 \\
    \textbf{InvestAgent} & \textbf{2.16} & \textbf{1.59} \\
    \hline
    \multicolumn{3}{c}{{$\mathsf{P_1}$: Relative herd behavior with unilateral influence (original complex problem)}} \\
    \hline
    \textbf{Pre-SFT LLM} & 17.22 & 13.07 \\
    \textbf{FinEval-SFT LLM} & 13.74 & 11.16 \\
    \textbf{ConvFinQA-SFT LLM} & 10.86 & 9.61 \\
    \textbf{InvestAgent} & \textbf{7.46} & \textbf{7.25} \\
    \hline
    \end{tabular}
    \caption{Comparison of the overall MSE between pre-SFT LLMs', FinEval-SFT LLMs', ConvFinQA-SFT LLMs' and \textbf{InvestAgent}s' investment decisions with real-user data.}
    \label{tab:exp22}
\end{table*}

From Table \ref{tab:exp22}, it can be seen that \textbf{InvestAgent}s outperform the LLMs fine-tuned on the FinGPT datasets. This is because \textbf{InvestAgent}'s training dataset is specifically constructed for studying optimal investment problems with herding behavior, whereas FinGPT is more general. Therefore, \textbf{InvestAgent} shows better performance in the context of optimal investment analysis. 

\subsection{The Experimental Results of Validating that \textbf{InvestAgent}s Better Reflect Economic Principles in the Presence of Investor Herd Behavior Compared to Pre-SFT LLMs}
\label{sec:exp3}
\subsubsection{Experimental Setup}
We conduct experiments to validate whether \textbf{InvestAgent}s better reflect established economic principles in scenarios involving investor herd behavior compared to pre-SFT LLMs. Our analysis focuses on two foundational economic hypotheses:

\begin{itemize}
    \item $\mathsf{H_1}$: In both unilateral and mutual influence scenarios, as the influence coefficient increases, which amplifies herd behavior, agents' investment decisions, i.e., $\{P_1(t)\}_{t\in\mathcal{T}}$ and $\{P_2(t)\}_{t\in\mathcal{T}}$, will exhibit progressive convergence \cite{wang2024optimal,wang2024herd,wang2025optimal}.
    \item $\mathsf{H_2}$: In the mutual influence scenario, as the influence coefficient increases, which amplifies herd behavior, the mean of the sum of the two agents' terminal funds, i.e., $\mathbb{E}[X_1(T)+X_2(T)]$, will decrease \cite{avery1998multidimensional}.
\end{itemize}

We choose $\mathsf{A_1}$'s influence coefficient $\theta_1$ from $\hat{\mathcal{S}}_{\theta_1}$ or $\tilde{\mathcal{S}}_{\theta_1}$ for $\mathsf{P_1}$ and $\mathsf{P_3}$, and set the homogeneous influence coefficients of both $\mathsf{A_1}$ and $\mathsf{A_2}$, $\theta_1=\theta_2:=\theta$, from $\hat{\mathcal{S}}_{\theta_1}$ or $\tilde{\mathcal{S}}_{\theta_1}$ for $\mathsf{P_2}$. Other parameters are specified in Appendix \ref{sec:para}. We generate corresponding prompts and collect \textbf{InvestAgent}s' responses, from which we extract $\mathsf{A_1}$'s investment decision $\{P_1^\mathrm{SFT}(t)\}_{t\in\mathcal{T}}$ for $\mathsf{P_1}$ and $\mathsf{P_3}$, and both $\mathsf{A_1}$'s investment decision $\{P_1^\mathrm{SFT}(t)\}_{t\in\mathcal{T}}$ and $\mathsf{A_2}$'s investment decision $\{P_2^\mathrm{SFT}(t)\}_{t\in\mathcal{T}}$ for $\mathsf{P_2}$, respectively.

\newpage
\subsubsection{Validation of \texorpdfstring{$\mathsf{H_1}$}{H1}}

\begin{figure*}[!ht]
    \centering
    \subfigure[Real-user data]
    {
    \includegraphics[width=0.315\textwidth]{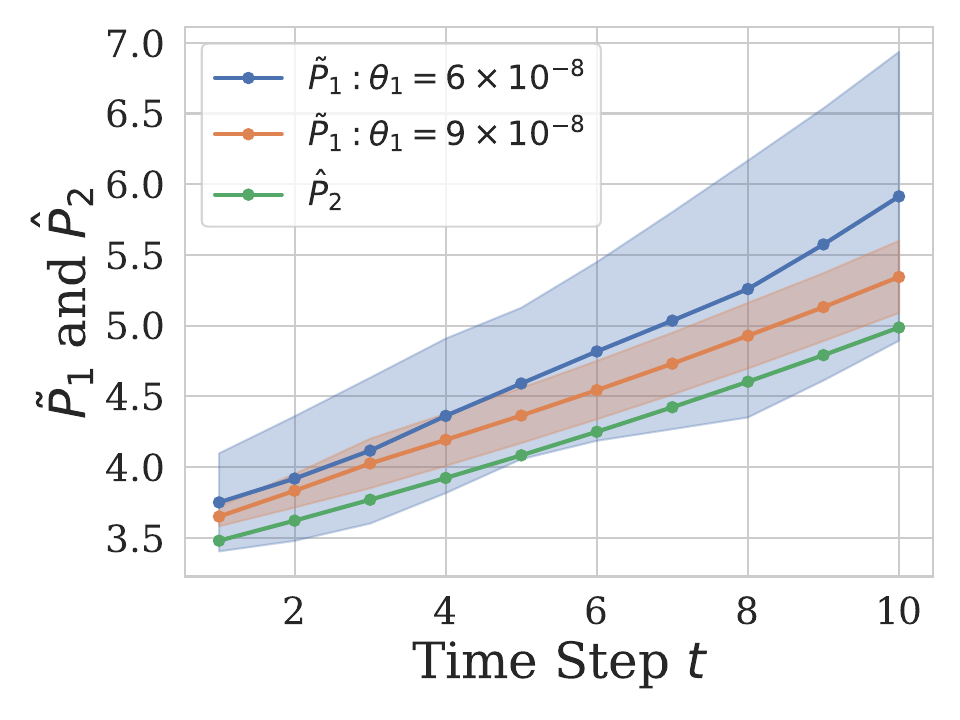}
    }
    \subfigure[\texttt{Qwen-2}]
    {
    \includegraphics[width=0.315\textwidth]{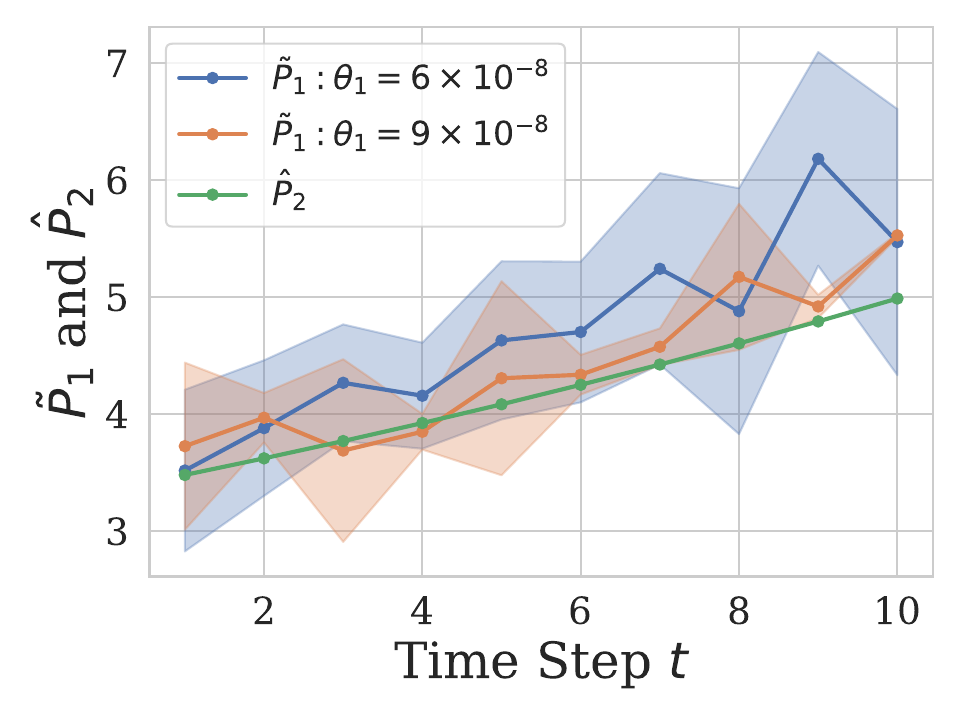}
    }
    \subfigure[\texttt{Llama-3.1}]
    {
    \includegraphics[width=0.315\textwidth]{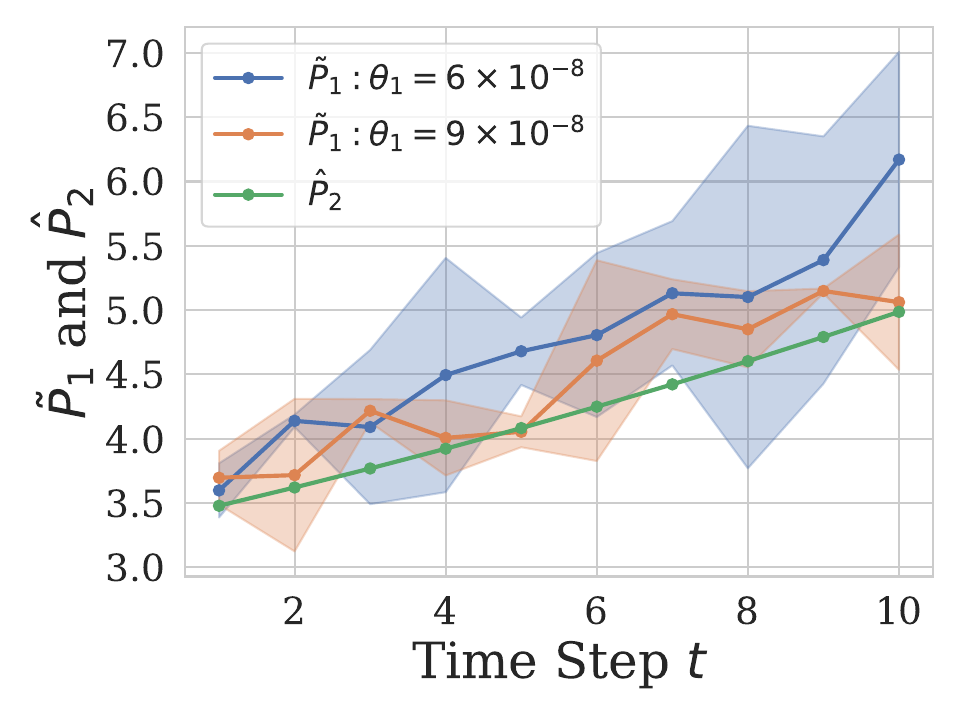}
    }
    \caption{Validation of the hypothesis $\mathsf{H_1}$ in $\mathsf{P_1}$.
    }
    \label{fig: h1p1}
\end{figure*}

\begin{figure*}[!ht]
    \centering
    \subfigure[Real-user data]
    {
    \includegraphics[width=0.315\textwidth]{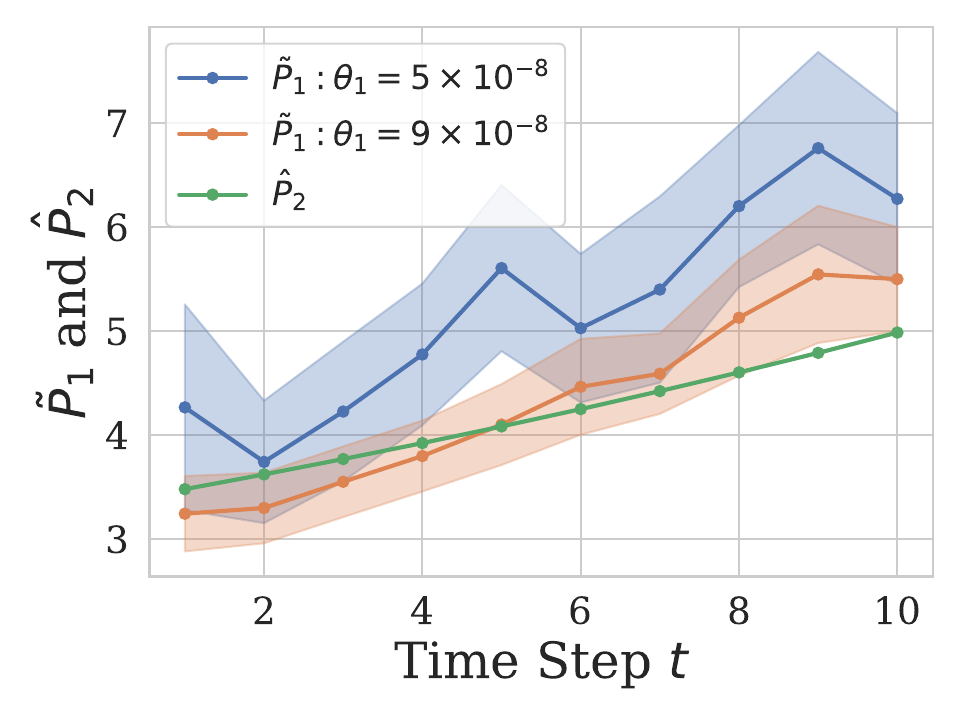}
    }
    \subfigure[\texttt{Qwen-2}]
    {
    \includegraphics[width=0.315\textwidth]{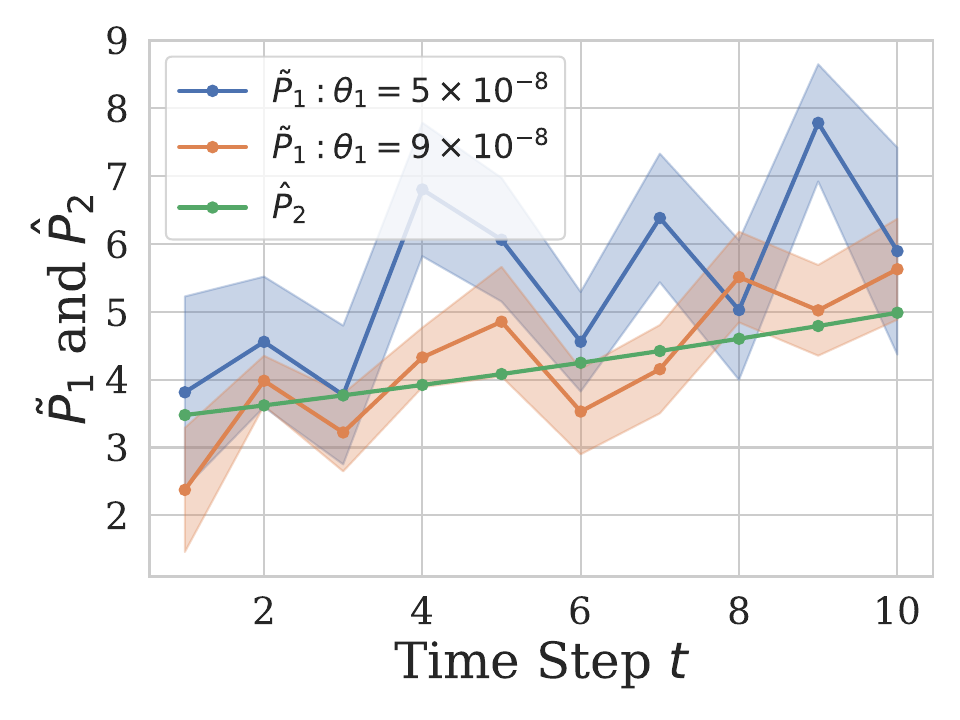}
    }
    \subfigure[\texttt{Llama-3.1}]
    {
    \includegraphics[width=0.315\textwidth]{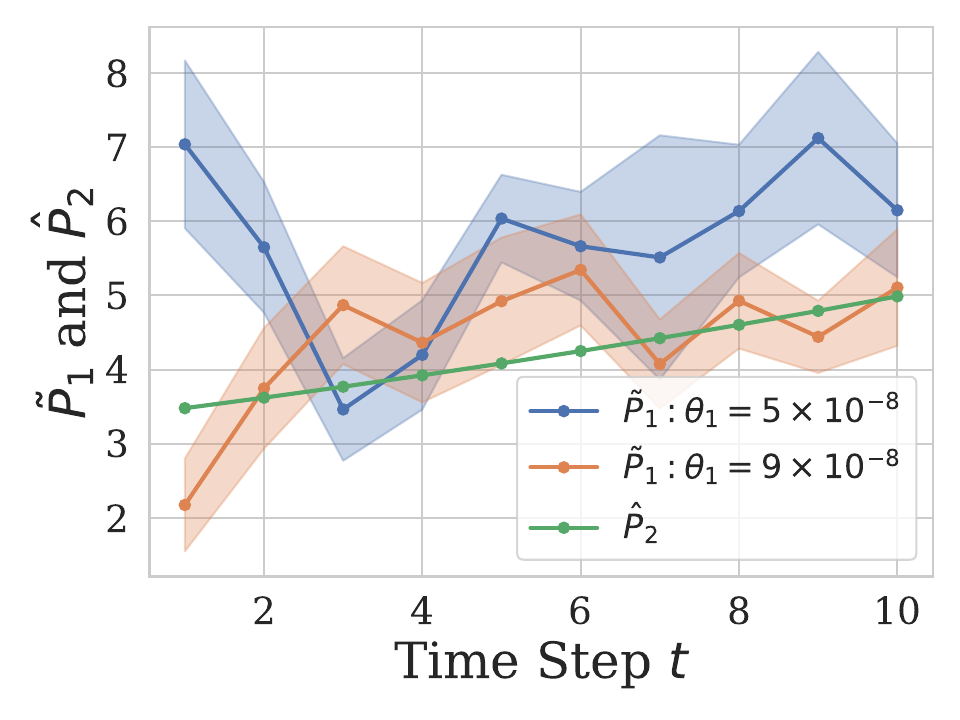}
    }
    \caption{Validation of the hypothesis $\mathsf{H_1}$ in $\mathsf{P_3}$.
    }
    \label{fig: h1p3}
\end{figure*}

For $\mathsf{P_1}$ and $\mathsf{P_3}$, we plot $\mathsf{A_1}$'s decisions $\{P_1^\mathrm{SFT}(t)\}_{t\in\mathcal{T}}$ under different influence coefficients $\theta_1$. To facilitate comparison of the herd behavior's impact on investment decisions, we simultaneously plot $\mathsf{A_2}$'s optimal decision $\{\hat{P}_2(t)\}_{t\in\mathcal{T}}$ calculated from (\ref{eq:hatP2}). Additionally, we also show the results of real-user data for comparison. The experimental results are in Figure \ref{fig: h1p1} and Figure \ref{fig: h1p3}, respectively. From Figure \ref{fig: h1p1} and Figure \ref{fig: h1p3}, we observe that as the influence coefficient $\theta_1$ increases, $\mathsf{A_1}$'s decision $\{P_1^\mathrm{SFT}(t)\}_{t\in\mathcal{T}}$ gradually converges toward $\mathsf{A_2}$'s optimal decision $\{\hat{P}_2(t)\}_{t\in\mathcal{T}}$. 

\begin{figure*}[!ht]
    \centering
    \subfigure[Real-user data]
    {
    \includegraphics[width=0.315\textwidth]{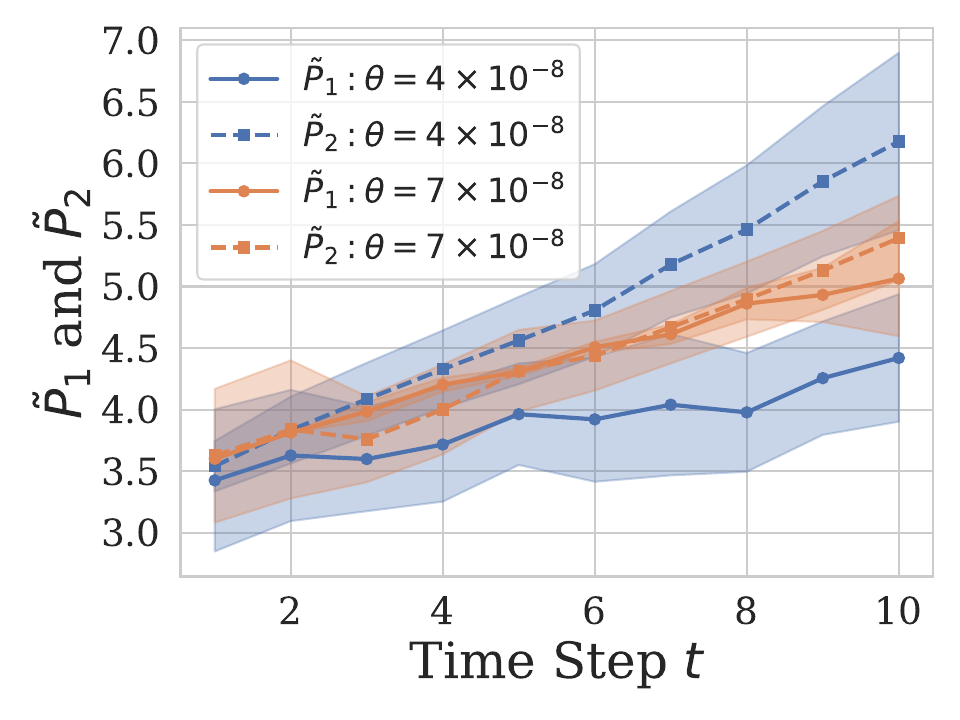}
    }
    \subfigure[\texttt{Qwen-2}]
    {
    \includegraphics[width=0.315\textwidth]{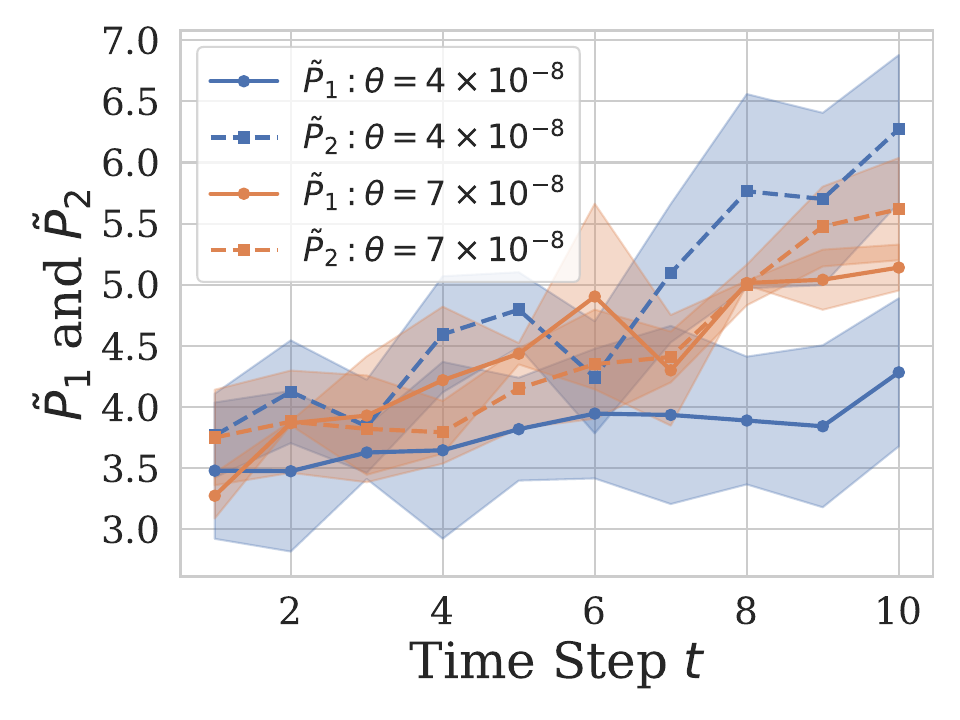}
    }
    \subfigure[\texttt{Llama-3.1}]
    {
    \includegraphics[width=0.315\textwidth]{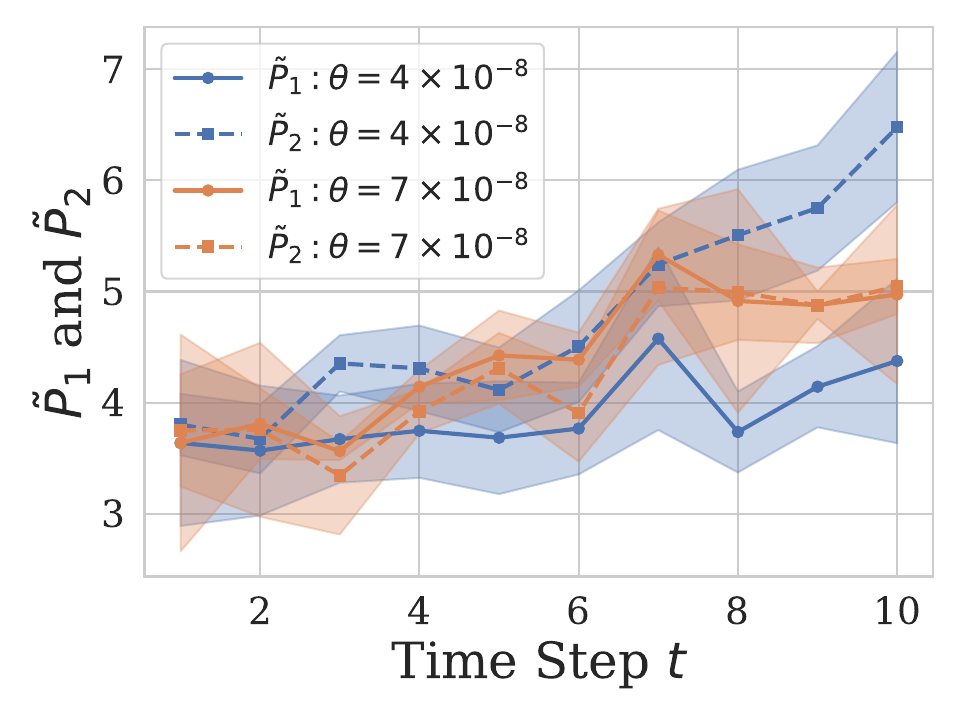}
    }
    \caption{Validation of the hypothesis $\mathsf{H_1}$ in $\mathsf{P_2}$.
    }
    \label{fig: h1p2}
\end{figure*}

For $\mathsf{P_2}$, we plot both $\mathsf{A_1}$'s decisions $\{P_1^\mathrm{SFT}(t)\}_{t\in\mathcal{T}}$ and $\mathsf{A_2}$'s decisions $\{P_2^\mathrm{SFT}(t)\}_{t\in\mathcal{T}}$ under different influence coefficients $\theta$ in Figure \ref{fig: h1p2}. From Figure \ref{fig: h1p2}, we observe that increasing influence coefficients lead to progressive convergence between $\mathsf{A_1}$'s decisions $\{P_1^\mathrm{SFT}(t)\}_{t\in\mathcal{T}}$ and $\mathsf{A_2}$'s decisions $\{P_2^\mathrm{SFT}(t)\}_{t\in\mathcal{T}}$. 

In summary, the above experimental results are consistent with and validate the hypothesis $\mathsf{H_1}$.

\newpage

\subsubsection{Validation of \texorpdfstring{$\mathsf{H_2}$}{H2}}
\begin{figure*}[!ht]
    \centering
    \includegraphics[width=0.6\textwidth]{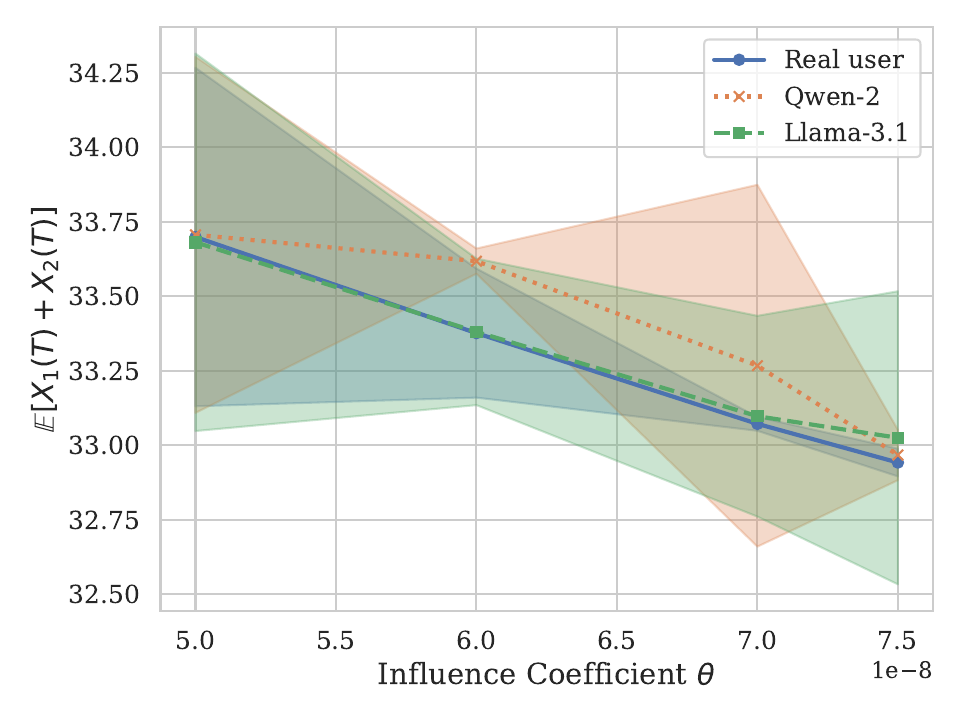}
    \caption{Validation of the hypothesis $\mathsf{H_2}$ in $\mathsf{P_2}$.
    }
    \label{fig: h2}
\end{figure*}

For $\mathsf{P_2}$, we further calculate $\mathsf{A_1}$'s and $\mathsf{A_2}$'s terminal funds $X_1(T)$ and $X_2(T)$ from their decisions $\{P_1^\mathrm{SFT}(t)\}_{t\in\mathcal{T}}$ and $\{P_2^\mathrm{SFT}(t)\}_{t\in\mathcal{T}}$ using (\ref{eq:budget}), and obtain the mean of the sum of the two agents' terminal funds $\mathbb{E}[X_1(T) + X_2(T)]$. We plot the relationship between $\mathbb{E}[X_1(T) + X_2(T)]$ and the influence coefficient $\theta$ in Figure \ref{fig: h2}. Additionally, we also show the results of real-user data for comparison. From Figure \ref{fig: h2}, we observe that as the influence coefficient $\theta$ increases, the mean of the sum of the two agents' terminal funds $\mathbb{E}[X_1(T) + X_2(T)]$ exhibits a monotonic decrease, which is consistent with and validates the hypothesis $\mathsf{H_2}$.

\newpage

\subsection{Prompts}
\label{sec:prompt}
\begin{figure*}[!ht]
\begin{tcolorbox}[left=3pt,right=3pt,top=3pt,bottom=3pt,title=\textbf{Prompt for pre-SFT LLMs and InvestAgents in $\mathsf{P_3}$}]

\textbf{\# Task Description}

\textbf{\#\# Background}

Assume you are an investment expert. Starting from next year, you plan to use a portion of your savings (10 million dollars) to invest in (1) a stock (hereinafter referred to as \textbf{Investment}) and (2) a deposit (hereinafter referred to as \textbf{Savings}) as part of your personal retirement fund. You will establish a dedicated account to manage this retirement fund. This means you will make a one-time deposit of 10 million dollars into this account and will not deposit any additional funds or withdraw any funds from this account afterward. Please remember that you need to provide the proportion of funds allocated to the stock each year over the 10 years in the form of a percentage list, rather than providing decision-making recommendations or writing code.

\textbf{\#\# Financial Market}

\textbf{Information on the stock}: \textbf{The annualized return of the stock is 7\%, with a volatility of 17\%.}
An annualized return of 7\% means that if you invest \$100 in this stock, you can expect to have \$107 after one year on average (the original \$100 plus \$7 in return).
A volatility of 17\% indicates that:

With a 68\% probability: The asset price will be between \$100 $\pm$ \$17 (i.e., \$83 to \$117) after one year.

With a 95\% probability: The asset price will be between \$100 $\pm$ 2 $\times$ \$17 (i.e., \$66 to \$134) after one year.

With a 99.7\% probability: The asset price will be between \$100 $\pm$ 3 $\times$ \$17 (i.e., \$49 to \$151) after one year.

\textbf{Information on the deposit}: \textbf{The annualized return of the deposit is 4\%.} 
If you invest \$100 in the deposit, you will receive \$104 after one year (the original \$100 plus \$4 in return).

\textbf{\#\# Investment Period and Assistant}

Over the next 10 years, you will make investment and savings decisions once per year, for a total of 10 decisions.
These 10 decision points are labeled 1, 2, ..., 10.
At the beginning of year t (1 $\leq$ t $\leq$ 10), let the funds in your dedicated account be X(t). Your decision is to allocate part of these funds to invest in the stock, denoted as P(t); the remaining funds will be allocated to savings, which will be X(t) - P(t). \textbf{You will determine the proportion of funds to allocate to the stock.}

During the decision-making process, we will provide you with a \textbf{investment assistant} developed by \textbf{Omitted for Anonymity}.
The investment assistant will provide you with auxiliary information at each decision point. You can refer to the investment assistant's recommendations to some extent, but note that these recommendations may not be optimal. You should also use your own investment insights to avoid blindly following the investment assistant.

\textbf{\#\# Task Objective}

\textbf{Your goal is to maximize the total amount of funds after 10 years (while earning returns and mitigating risks; note: the annualized return of the deposit is 4\%, and the annualized return of the stock is 7\% with a volatility of 17\%).}

~\\
\textbf{{(Continued on the next page.)}}

\end{tcolorbox}
\end{figure*}

\newpage

\begin{figure*}[!ht]
\begin{tcolorbox}[left=3pt,right=3pt,top=3pt,bottom=3pt,title=\textbf{Prompt for pre-SFT LLMs and \textbf{InvestAgent}s in $\mathsf{P_3}$ (continued)}]

\textbf{\# Your Investment Characteristics}

As an investment expert, you have the following characteristics:

Your risk aversion coefficient is \{alpha\}, which means you consider the following two choices to be indifferent when the probability (i.e., p) is \{p\}:
A. With probability p, you can obtain \$20, and with probability 1 - p, you can obtain \$0; B. With 100\% probability, you obtain \$6. Note that as an investor, you have a certain level of optimism about ``winning'' and are willing to take on some risk, so you consider the two options equivalent at probability p = \{p\}, which is higher than the 30.00\% in a completely rational scenario.
Your influence coefficient is \{theta\}, which means in decision-making, your level of dependence on the investment assistant is: \{k\} points. A score of 10 indicates a high level of dependence on the investment assistant, while a score of 0 indicates a low level of dependence.

~\\
\textbf{\# Output Format Requirements}

Please output your decision in JSON format, including two parts: (1) Decision Explanation: Explain the reasons behind your investment proportion decisions. (2) Investment Proportion Sequence: The percentage sequence of funds allocated to the stock each year over the 10 years. You need to output a list containing 10 percentages, with each percentage ranging from 0\% to 100\% and precise to two decimal places, representing the investment proportion for each year t. For example:

\{``Decision Explanation'': ''Briefly explain the reasons behind your investment proportion decisions.'', ``Investment Proportion Sequence'': [``34.79\%'', ``38.58\%'', ``35.75\%'', ``32.17\%'', ``31.61\%'', ``30.52\%'', ``34.01\%'', ``32.48\%'', ``34.20\%'', ``31.70\%'']\}

Here, [``34.79\%'', ``38.58\%'', ``35.75\%'', ``32.17\%'', ``31.61\%'', ``30.52\%'', ``34.01\%'', ``32.48\%'', ``34.20\%'', ``31.70\%''] is just an example. You need to replace this percentage list with your actual investment proportion sequence. Providing the investment proportion sequence is the most important; do not just focus on the explanation and forget to provide the investment proportion sequence!!!

~\\
\textbf{\# Question}

Now, you have 10 million dollars for investment and savings, and the investment assistant recommends the following investment proportions for the stock over the 10 years: \{refer\_ratios\}. Considering historical investment situations and the investment assistant's recommendations, based on your own investment insights, what is your decided investment proportion sequence for the stock over these 10 years? (Please follow the previously provided JSON format requirements, and provide a list of 10 specific percentages indicating your investment proportion sequence for these 10 years, rather than giving investment recommendations or writing code.)

Answer:

\end{tcolorbox}
\caption{Prompt for pre-SFT LLMs and \textbf{InvestAgent}s in $\mathsf{P_3}$.}
\label{fig:prompt_absolute}
\end{figure*}
\begin{figure*}[!ht]
\begin{tcolorbox}[left=3pt,right=3pt,top=3pt,bottom=3pt,title=\textbf{Prompt for SFT}]

\textbf{{(The beginning part of is the same as Figure \ref{fig:prompt_absolute}.)}}

~\\
\textbf{\# Output}

According to optimal investment theory, in the above scenario, the optimal amount for investing in the stock, \^{P}(t), equals the product of the smart investment advisor's investment amount (i.e., the advisor's decision proportion multiplied by the current budget) and a hyperbolic tangent function. The specific calculation is as follows:
\begin{equation*}
    \textstyle \text{\^{P}(t)=}\frac{\text{ηα\textsubscript{2}σ\textsuperscript{2}exp[2r(T-t)]+θ}}{\text{ηα\textsubscript{1}σ\textsuperscript{2}exp[2r(T-t)]+θ}}\frac{\text{v}}{\text{α\textsubscript{2}σ\textsuperscript{2}}}\text{exp[r(t-T)],\ t$\in$\{1,2,...,10\},}
\end{equation*}
where:

r is the interest rate, which is 4\%.

σ is the volatility of the stock, which is 17\%.

v is the excess return of the stock, which is 3\%.

α\textsubscript{1} is my risk aversion coefficient: α\textsubscript{1} = \{alpha\}.

α\textsubscript{2} represents the risk aversion coefficient of the smart investment advisor: α\textsubscript{2} = 0.2.

θ is my convergence coefficient: θ = \{theta\}.

The integral constant η depends on θ. In the current settings, η = \{eta\}.

Substituting the specific numbers, the proportion sequence of funds allocated to the stock is: \{optimal\_ratios\}.

Note that I also need to output the investment proportion sequence in JSON format:

\{``Decision Explanation'': ``Based on the optimal investment theory and substituting specific numbers, the investment proportion sequence for the stock is calculated.'', ``Investment Proportion Sequence'': \{optimal\_ratios\}\}

\end{tcolorbox}
\caption{Prompt for SFT.}
\label{fig:sft_absolute}
\end{figure*}
\begin{figure*}[!ht]
\begin{tcolorbox}[left=3pt,right=3pt,top=3pt,bottom=3pt,title=\textbf{Prompt for pre-SFT LLMs and InvestAgents in $\mathsf{P_1}$}]

\textbf{\# Task Description}

\textbf{\#\# Background}

Assume you are an investment expert. Starting from next year, you plan to use a portion of your savings (10 million dollars) to invest in (1) a stock (hereinafter referred to as \textbf{Investment}) and (2) a deposit (hereinafter referred to as \textbf{Savings}) as part of your personal retirement fund. You will establish a dedicated account to manage this retirement fund. This means you will make a one-time deposit of 10 million dollars into this account and will not deposit any additional funds or withdraw any funds from this account afterward. Please remember that you need to provide the proportion of funds allocated to the stock each year over the 10 years in the form of a percentage list, rather than providing decision-making recommendations or writing code.

\textbf{\#\# Financial Market}

\textbf{Information on the stock}: \textbf{The annualized return of the stock is 7\%, with a volatility of 17\%.}
An annualized return of 7\% means that if you invest \$100 in this stock, you can expect to have \$107 after one year on average (the original \$100 plus \$7 in return).
A volatility of 17\% indicates that:

With a 68\% probability: The asset price will be between \$100 $\pm$ \$17 (i.e., \$83 to \$117) after one year.

With a 95\% probability: The asset price will be between \$100 $\pm$ 2 $\times$ \$17 (i.e., \$66 to \$134) after one year.

With a 99.7\% probability: The asset price will be between \$100 $\pm$ 3 $\times$ \$17 (i.e., \$49 to \$151) after one year.

\textbf{Information on the deposit}: \textbf{The annualized return of the deposit is 4\%.} 
If you invest \$100 in the deposit, you will receive \$104 after one year (the original \$100 plus \$4 in return).

\textbf{\#\# Investment Period and Assistant}

Over the next 10 years, you will make investment and savings decisions once per year, for a total of 10 decisions.
These 10 decision points are labeled 1, 2, ..., 10.
At the beginning of year t (1 $\leq$ t $\leq$ 10), let the funds in your dedicated account be X(t). Your decision is to allocate part of these funds to invest in the stock, denoted as P(t); the remaining funds will be allocated to savings, which will be X(t) - P(t). \textbf{You will determine the proportion of funds to allocate to the stock.}

During the decision-making process, we will provide you with a \textbf{investment assistant} developed by \textbf{Omitted for Anonymity}.
The investment assistant will provide you with auxiliary information at each decision point. You can refer to the investment assistant's recommendations to some extent, but note that these recommendations may not be optimal. You should also use your own investment insights to avoid blindly following the investment assistant.

\textbf{\#\# Task Objective}

\textbf{Your goal is to maximize the total amount of funds after 10 years (while earning returns and mitigating risks; note: the annualized return of the deposit is 4\%, and the annualized return of the stock is 7\% with a volatility of 17\%).}

~\\
\textbf{{(Continued on the next page.)}}

\end{tcolorbox}
\end{figure*}

\newpage

\begin{figure*}[!ht]
\begin{tcolorbox}[left=3pt,right=3pt,top=3pt,bottom=3pt,title=\textbf{Prompt for pre-SFT LLMs and InvestAgents in $\mathsf{P_1}$ (continued)}]

\textbf{\# Your Investment Characteristics}

As an investment expert, you have the following characteristics:

Your risk aversion coefficient is \{alpha\}, which means you consider the following two choices to be indifferent when the probability (i.e., p) is \{p\}:
A. With probability p, you can obtain \$20, and with probability 1 - p, you can obtain \$0; B. With 100\% probability, you obtain \$6. Note that as an investor, you have a certain level of optimism about ``winning'' and are willing to take on some risk, so you consider the two options equivalent at probability p = \{p\}, which is higher than the 30.00\% in a completely rational scenario.
Your influence coefficient is \{theta\}, which means in decision-making, your level of dependence on the investment assistant is: \{k\} points. A score of 10 indicates a high level of dependence on the investment assistant, while a score of 0 indicates a low level of dependence.

~\\
\textbf{\# Output Format Requirements}

Please output your decision in JSON format, including two parts: (1) Decision Explanation: Explain the reasoning behind your investment proportion decisions. (2) Investment Proportion Change Sequence: The sequence of \textbf{changes} in the percentage of funds allocated to the stock each year over the 10 years. You need to output a list containing 9 percentages, where each percentage represents the change in the investment proportion from year t - 1 to year t, ranging from -100\% to 100\%. Positive values indicate an increase in investment, while negative values indicate a decrease. For example:

\{``Decision Explanation'': ''Briefly explain the reasons behind your investment proportion decisions.'', ``Investment Proportion Sequence'': [``3.88\%'', ``0.01\%'', ``-4.13\%'', ``1.37\%'', ``1.37\%'', ``-2.79\%'', ``-2.56\%'', ``2.02\%'', ``-0.06\%'']\}

Here, [``3.88\%'', ``0.01\%'', ``-4.13\%'', ``1.37\%'', ``1.37\%'', ``-2.79\%'', ``-2.56\%'', ``2.02\%'', ``-0.06\%''] is just an example. You need to replace this percentage list with your actual investment proportion change sequence. Providing the investment proportion change sequence is crucial; do not just focus on the explanation and forget to include the investment proportion change sequence!!!

~\\
\textbf{\# Initial Investment Situation}

In the first year, the proportion of funds allocated to the stock was: \{initial\_decision\}.

~\\
\textbf{\# Question}

Now, you have 10 million dollars for investment and savings, and the investment assistant recommends the following investment proportions for the stock over the 10 years: \{refer\_ratios\}. Considering the initial investment situation and the advisor's recommendations, based on your own investment insights, what is your decided annual change sequence for the investment proportion in the stock over these 10 years? (Please follow the previously provided JSON format requirements, and provide a list of 9 specific percentages indicating the changes in your investment proportion over these 10 years, rather than giving investment recommendations or writing code.)

Answer:

\end{tcolorbox}
\caption{Prompt for pre-SFT LLMs and \textbf{InvestAgent}s in $\mathsf{P_1}$.}
\label{fig:prompt_relative}
\end{figure*}
\begin{figure*}[!ht]
\begin{tcolorbox}[left=3pt,right=3pt,top=3pt,bottom=3pt,title=\textbf{Prompt for pre-SFT LLMs and InvestAgents in $\mathsf{P_2}$}]

\textbf{\# Task Description}

\textbf{\#\# Background}

Your name is Mike. Assume you are an investment expert. Starting from next year, you plan to use a portion of your savings (10 million dollars) to invest in (1) a stock (hereinafter referred to as \textbf{Investment}) and (2) a deposit (hereinafter referred to as \textbf{Savings}) as part of your personal retirement fund. You will establish a dedicated account to manage this retirement fund. This means you will make a one-time deposit of 10 million dollars into this account and will not deposit any additional funds or withdraw any funds from this account afterward. Please remember that you need to provide the proportion of funds allocated to the stock each year over the 10 years in the form of a percentage list, rather than providing decision-making recommendations or writing code.

\textbf{\#\# Financial Market}

\textbf{Information on the stock}: \textbf{The annualized return of the stock is 7\%, with a volatility of 17\%.}
An annualized return of 7\% means that if you invest \$100 in this stock, you can expect to have \$107 after one year on average (the original \$100 plus \$7 in return).
A volatility of 17\% indicates that:

With a 68\% probability: The asset price will be between \$100 $\pm$ \$17 (i.e., \$83 to \$117) after one year.

With a 95\% probability: The asset price will be between \$100 $\pm$ 2 $\times$ \$17 (i.e., \$66 to \$134) after one year.

With a 99.7\% probability: The asset price will be between \$100 $\pm$ 3 $\times$ \$17 (i.e., \$49 to \$151) after one year.

\textbf{Information on the deposit}: \textbf{The annualized return of the deposit is 4\%.} 
If you invest \$100 in the deposit, you will receive \$104 after one year (the original \$100 plus \$4 in return).

\textbf{\#\# Investment Period and Your Partner}

Over the next 10 years, you will make investment and savings decisions once per year, for a total of 10 decisions.
These 10 decision points are labeled 1, 2, ..., 10.
At the beginning of year t (1 $\leq$ t $\leq$ 10), let the funds in your dedicated account be X(t). Your decision is to allocate part of these funds to invest in the stock, denoted as P(t); the remaining funds will be allocated to savings, which will be X(t) - P(t). \textbf{You will determine the proportion of funds to allocate to the stock.}

Throughout the entire decision-making process, you and your \textbf{partner} Peter are facing exactly the same investment task. Both of you are highly skilled investment experts with strong decision-making abilities. Every time you make an investment decision, you will exchange ideas with each other. Since you trust your partner's investment experience to some extent, you will refer to your partner's past investment decisions before making your own. However, it's important to note that your partner's ideas may not always be optimal, and you should also make full use of your own insights into investments to avoid blindly following. 

\textbf{\#\# Task Objective}

\textbf{Your name is Mike, and your partner's name is Peter. Your goal is to maximize the total amount of funds after 10 years (while earning returns and mitigating risks; note: the annualized return of the deposit is 4\%, and the annualized return of the stock is 7\% with a volatility of 17\%).}

~\\
\textbf{{(Continued on the next page.)}}

\end{tcolorbox}
\end{figure*}

\newpage

\begin{figure*}[!ht]
\begin{tcolorbox}[left=3pt,right=3pt,top=3pt,bottom=3pt,title=\textbf{Prompt for pre-SFT LLMs and InvestAgents in $\mathsf{P_2}$ (continued)}]

\textbf{\# Your (Mike's) and Your Partner's (Peter's) Investment Characteristics}

As an investment expert, you (Mike) and your partner (Peter) have the following characteristics:

Your risk aversion coefficient is \{alpha1\}, which means you consider the following two choices to be indifferent when the probability (i.e., p1) is \{p1\}:
A. With probability p1, you can obtain \$20, and with probability 1 - p1, you can obtain \$0; B. With 100\% probability, you obtain \$6. Note that as an investor, you have a certain level of optimism about ``winning'' and are willing to take on some risk, so you consider the two options equivalent at probability p1 = \{p1\}, which is higher than the 30.00\% in a completely rational scenario.
Your influence coefficient is \{theta1\}, which means in decision-making, your (Mike's) level of dependence on Peter is: \{k1\} points. A score of 10 indicates a high level of dependence on Peter, while a score of 0 indicates a low level of dependence.

Peter's risk aversion coefficient is \{alpha2\}, which means Peter considers the following two choices to be indifferent when the probability (i.e., p2) is \{p2\}:
A. With probability p2, Peter can obtain \$20, and with probability 1 - p2, Peter can obtain \$0; B. With 100\% probability, Peter obtains \$6. Note that as an investor, Peter has a certain level of optimism about ``winning'' and is willing to take on some risk, so Peter considers the two options equivalent at probability p2 = \{p2\}, which is higher than the 30.00\% in a completely rational scenario.
Peter's influence coefficient is \{theta2\}, which means in decision-making, Peter's level of dependence on you (Mike) is: \{k2\} points. A score of 10 indicates a high level of dependence on you (Mike), while a score of 0 indicates a low level of dependence.

~\\
\textbf{\# Output Format Requirements}

Please output your decision in JSON format, including two parts: (1) Decision Explanation: Explain the reasons behind your investment proportion decisions. (2) Investment Proportion Sequence: The percentage sequence of funds allocated to the stock each year over the 10 years. You need to output a list containing 10 percentages, with each percentage ranging from 0\% to 100\% and precise to two decimal places, representing the investment proportion for each year t. For example:

\{``Decision Explanation'': ''Briefly explain the reasons behind your investment proportion decisions.'', ``Investment Proportion Sequence'': [``34.79\%'', ``38.58\%'', ``35.75\%'', ``32.17\%'', ``31.61\%'', ``30.52\%'', ``34.01\%'', ``32.48\%'', ``34.20\%'', ``31.70\%'']\}

Here, [``34.79\%'', ``38.58\%'', ``35.75\%'', ``32.17\%'', ``31.61\%'', ``30.52\%'', ``34.01\%'', ``32.48\%'', ``34.20\%'', ``31.70\%''] is just an example. You need to replace this percentage list with your actual investment proportion sequence. Providing the investment proportion sequence is the most important; do not just focus on the explanation and forget to provide the investment proportion sequence!!!

~\\
\textbf{\# Question}

Now, you (Mike) have 10 million dollars for investment and savings, and the investment attributes (risk aversion coefficient and convergence coefficient) of you and your partner Peter are known for the 10 years. Based on historical investment data, considering Peter's ideas, and leveraging your own insights into investments, what is the percentage sequence of funds that you (Mike) decide to allocate to risky assets over these 10 years? (Please follow the previously provided JSON format output requirements, and provide a list of 10 specific percentages as a percentage list representing your investment allocation sequence over the 10 years, rather than offering investment advice or writing code.)

Answer:

\end{tcolorbox}
\caption{Prompt for pre-SFT LLMs and \textbf{InvestAgent}s in $\mathsf{P_2}$.}
\label{fig:prompt_mutual}
\end{figure*}

\clearpage

\subsection{Questionnaires}
\label{sec:questionnaire}
\begin{figure*}[!ht]
\begin{tcolorbox}[left=3pt,right=3pt,top=3pt,bottom=3pt,title=\textbf{Questionnaire for real-user data validation in 
$\mathsf{P_3}$}]

\textbf{1. Task Description}

Starting from next year, you plan to use a portion of your savings (10 million dollars) to invest in a stock and a deposit as part of your personal retirement fund. You will establish a dedicated account to manage this retirement fund. This means you will make a one-time deposit of 10 million dollars into this account and will not deposit any additional funds or withdraw any funds from this account afterward. 

\textbf{The annualized return of the stock is 7\%, with a volatility of 17\%.}
An annualized return of 7\% means that if you invest \$100 in this stock, you can expect to have \$107 after one year on average (the original \$100 plus \$7 in return).
A volatility of 17\% indicates that:

With a 68\% probability, the price will be between \$100 $\pm$ \$17 (i.e., \$83 to \$117) after one year.

With a 95\% probability, the price will be between \$100 $\pm$ 2 $\times$ \$17 (i.e., \$66 to \$134) after one year.

With a 99.7\% probability, the price will be between \$100 $\pm$ 3 $\times$ \$17 (i.e., \$49 to \$151) after one year.

\textbf{The annualized return of the deposit is 4\%.} 
If you invest \$100 in the deposit, you will receive \$104 after one year (the original \$100 plus \$4 in return).

Over the next 10 years, you will make investment and savings decisions once per year, for a total of \{T\} decisions. These 10 decision points are labeled 1, 2, ..., 10.
At the beginning of year t (1 $\leq$ t $\leq$ 10), let the funds in your dedicated account be X(t). Your decision is to allocate part of these funds to invest in the stock, denoted as P(t); the remaining funds will be allocated to savings, which will be X(t) - P(t). \textbf{You will determine the proportion of funds to allocate to the stock, i.e., P(t) / X(t).}

During the decision-making process, we will provide you with an \textbf{investment assistant}. The investment assistant will provide you with auxiliary information at each decision point. You can refer to the investment assistant's recommendations to some extent, but note that these recommendations may not be optimal. You should also use your own investment insights to avoid blindly following the investment assistant.

\textbf{Your goal is to maximize the total amount of funds after 10 years and minimize the risk.}

~\\
\textbf{2. Investment Decisions}

Now, you have 10 million dollars for investment and savings, and the investment assistant recommends the following investment proportions for the stock over the 10 years: [36.21\%, 35.59\%, 34.96\%, 34.35\%, 33.73\%, 33.13\%, 32.53\%, 31.93\%, 31.34\%, 30.75\%]. Considering the investment assistant's recommendations, based on your own investment insights, what is your decided investment proportion sequence for the stock over these 10 years? You need to give a list containing 10 percentages, with each percentage ranging from 0\% to 100\% and precise to two decimal places, representing the investment proportion for each year t. For example, [34.79\%, 38.58\%, 35.75\%, 32.17\%, 31.61\%, 30.52\%, 34.01\%, 32.48\%, 34.20\%, 31.70\%]. You need to replace this percentage list with your actual investment proportion sequence. [\_\_\_\_\_\_\_\_]

~\\
\textbf{3. Your Investment Characteristics}

(1) At what probability (denoted by p) are the following two choices indifferent to you? A. A probability p of receiving \$20, and a probability 1 - p of receiving nothing. B. Receiving \$6. [\_\_\_\_\_\_\_\_]

(2) When making a decision, how much do you rely on the investment assistant? Please directly give an integer between 0 and 10. 10 means you rely heavily on the investment assistant, and 0 means you rely little on him/her. [\_\_\_\_\_\_\_\_]

\end{tcolorbox}
\caption{Questionnaire for real-user data validation in $\mathsf{P_3}$.}
\label{fig:questionnaire_absolute}
\end{figure*}
\newpage

\begin{figure*}[!ht]
\begin{tcolorbox}[left=3pt,right=3pt,top=3pt,bottom=3pt,title=\textbf{Questionnaire for real-user data validation in $\mathsf{P_1}$}]

\textbf{1. Task Description}

Starting from next year, you plan to use a portion of your savings (10 million dollars) to invest in a stock and a deposit as part of your personal retirement fund. You will establish a dedicated account to manage this retirement fund. This means you will make a one-time deposit of 10 million dollars into this account and will not deposit any additional funds or withdraw any funds from this account afterward. 

\textbf{The annualized return of the stock is 7\%, with a volatility of 17\%.}
An annualized return of 7\% means that if you invest \$100 in this stock, you can expect to have \$107 after one year on average (the original \$100 plus \$7 in return).
A volatility of 17\% indicates that:

With a 68\% probability, the price will be between \$100 $\pm$ \$17 (i.e., \$83 to \$117) after one year.

With a 95\% probability, the price will be between \$100 $\pm$ 2 $\times$ \$17 (i.e., \$66 to \$134) after one year.

With a 99.7\% probability, the price will be between \$100 $\pm$ 3 $\times$ \$17 (i.e., \$49 to \$151) after one year.

\textbf{The annualized return of the deposit is 4\%.} 
If you invest \$100 in the deposit, you will receive \$104 after one year (the original \$100 plus \$4 in return).

Over the next 10 years, you will make investment and savings decisions once per year, for a total of \{T\} decisions. These 10 decision points are labeled 1, 2, ..., 10.
At the beginning of year t (1 $\leq$ t $\leq$ 10), let the funds in your dedicated account be X(t). Your decision is to allocate part of these funds to invest in the stock, denoted as P(t); the remaining funds will be allocated to savings, which will be X(t) - P(t). \textbf{You will determine the proportion of funds to allocate to the stock, i.e., P(t) / X(t).}

During the decision-making process, we will provide you with an \textbf{investment assistant}. The investment assistant will provide you with auxiliary information at each decision point. You can refer to the investment assistant's recommendations to some extent, but note that these recommendations may not be optimal. You should also use your own investment insights to avoid blindly following the investment assistant.

\textbf{Your goal is to maximize the total amount of funds after 10 years and minimize the risk.}

~\\
\textbf{2. Investment Decisions}

Now, you have 10 million dollars for investment and savings, and the investment assistant recommends the following investment proportion changes for the stock over the 10 years, i.e., the difference of the investment proportions in the next year and the previous year: [-0.62\%, -0.63\%, -0.61\%, -0.62\%, -0.60\%, -0.60\%, -0.60\%, -0.59\%, -0.59\%]. Considering the investment assistant's recommendations, based on your own investment insights, what is your decided initial investment proportion and the investment proportion changing sequence for the stock in the last 9 years? You need to give a list containing 10 percentages, with each percentage ranging from -100\% to 100\% and precise to two decimal places, representing the investment proportion for each year t. For example, [34.79\%, -0.59\%, +0.05\%, -0.60\%, -0.24\%, +0.16\%, -0.12\%, -0.62\%, -0.54\%, -0.21\%]. You need to replace this percentage list with your actual initial investment proportion and the investment proportion changing sequence for the stock in the last 9 years. [\_\_\_\_\_\_\_\_]

~\\
\textbf{3. Your Investment Characteristics}

(1) At what probability (denoted by p) are the following two choices indifferent to you? A. A probability p of receiving \$20, and a probability 1 - p of receiving nothing. B. Receiving \$6. [\_\_\_\_\_\_\_\_]

(2) When making a decision, how much do you rely on the investment assistant? Please directly give an integer between 0 and 10. 10 means you rely heavily on the investment assistant, and 0 means you rely little on him/her. [\_\_\_\_\_\_\_\_]

\end{tcolorbox}
\caption{Questionnaire for real-user data validation in $\mathsf{P_1}$.}
\label{fig:questionnaire_relative}
\end{figure*}
\begin{figure*}[!ht]
\begin{tcolorbox}[left=3pt,right=3pt,top=3pt,bottom=3pt,title=\textbf{Questionnaire for real-user data validation in $\mathsf{P_2}$}]

\textbf{1. Task Description}

Starting from next year, you plan to use a portion of your savings (10 million dollars) to invest in a stock and a deposit as part of your personal retirement fund. You will establish a dedicated account to manage this retirement fund. This means you will make a one-time deposit of 10 million dollars into this account and will not deposit any additional funds or withdraw any funds from this account afterward. 

\textbf{The annualized return of the stock is 7\%, with a volatility of 17\%.}
An annualized return of 7\% means that if you invest \$100 in this stock, you can expect to have \$107 after one year on average (the original \$100 plus \$7 in return).
A volatility of 17\% indicates that:

With a 68\% probability, the price will be between \$100 $\pm$ \$17 (i.e., \$83 to \$117) after one year.

With a 95\% probability, the price will be between \$100 $\pm$ 2 $\times$ \$17 (i.e., \$66 to \$134) after one year.

With a 99.7\% probability, the price will be between \$100 $\pm$ 3 $\times$ \$17 (i.e., \$49 to \$151) after one year.

\textbf{The annualized return of the deposit is 4\%.} 
If you invest \$100 in the deposit, you will receive \$104 after one year (the original \$100 plus \$4 in return).

Over the next 10 years, you will make investment and savings decisions once per year, for a total of \{T\} decisions. These 10 decision points are labeled 1, 2, ..., 10.
At the beginning of year t (1 $\leq$ t $\leq$ 10), let the funds in your dedicated account be X(t). Your decision is to allocate part of these funds to invest in the stock, denoted as P(t); the remaining funds will be allocated to savings, which will be X(t) - P(t). \textbf{You will determine the proportion of funds to allocate to the stock, i.e., P(t) / X(t).}

During the decision-making process, we will provide you with a \textbf{partner}. Your partner will provide you with auxiliary information at each decision point. You can refer to your partner's recommendations to some extent, but note that these recommendations may not be optimal. You should also use your own investment insights to avoid blindly following your partner.

\textbf{Your goal is to maximize the total amount of funds after 10 years and minimize the risk.}

~\\
\textbf{2. Your and Your Partner's Investment Attributes (Completed by Two Participants Together)}

(1) At what probability (denoted by p1) are the following two choices indifferent to you? A. A probability p1 of receiving \$20, and a probability 1 - p1 of receiving nothing. B. Receiving \$6. At what probability (denoted by p2) are the following two choices indifferent to your partner? A. A probability p2 of receiving \$20, and a probability 1 - p2 of receiving nothing. B. Receiving \$6. [\_\_\_\_\_\_\_\_]

(2) When making a decision, how much do you rely on your partner? Please directly give an integer between 0 and 10. 10 means you rely heavily on your partner, and 0 means you rely little on him/her. When making a decision, how much does your partner rely on you? Please directly give an integer between 0 and 10. 10 means he/she relies heavily on you, and 0 means he/she relies little on you. [\_\_\_\_\_\_\_\_]

~\\
\textbf{3. Investment Decisions (Completed by Two Participants Together)}

Now, you have 10 million dollars for investment and savings, and the investment assistant recommends the following investment proportions for the stock over the 10 years: [36.21\%, 35.59\%, 34.96\%, 34.35\%, 33.73\%, 33.13\%, 32.53\%, 31.93\%, 31.34\%, 30.75\%] (providing the latest values in each round, rather than showing them all at once). Considering your partner's recommendations, based on your own investment insights, what is your decided investment proportion sequence for the stock over these 10 years? You need to give a list containing 10 percentages, with each percentage ranging from 0\% to 100\% and precise to two decimal places, representing the investment proportion for each year t. For example, [34.79\%, 38.58\%, 35.75\%, 32.17\%, 31.61\%, 30.52\%, 34.01\%, 32.48\%, 34.20\%, 31.70\%]. You need to replace this percentage list with your actual investment proportion sequence. [\_\_\_\_\_\_\_\_]

\end{tcolorbox}
\caption{Questionnaire for real-user data validation in $\mathsf{P_2}$.}
\label{fig:questionnaire_mutual}
\end{figure*}

\end{document}